\documentclass{article}
\usepackage[square,numbers]{natbib}
\usepackage[utf8]{inputenc} % allow utf-8 input
\usepackage[T1]{fontenc} % use 8-bit T1 fonts
\usepackage{url} % simple URL typesetting
\usepackage{booktabs} % professional-quality tables
\usepackage{amsfonts} % blackboard math symbols
\usepackage{nicefrac} % compact symbols for 1/2, etc.
\usepackage{microtype} % microtypography
\usepackage{lipsum}
\usepackage{fancyhdr} % header
\usepackage{graphicx} % graphics
\graphicspath{{media/}} % organize your images and other figures under media/ folder
\usepackage{xcolor}

\usepackage[para]{threeparttable}
\usepackage{amsmath}
\usepackage{nccmath}
\usepackage{tabularx}
\usepackage{multirow}
\usepackage{amssymb}
\usepackage{multicol}
\usepackage{array}
\usepackage{physics}
\usepackage{authblk}
\usepackage{float}
\usepackage{makecell}
\usepackage{palatino}
\usepackage{graphicx}
\graphicspath{ {} }
\usepackage[labelfont=bf]{caption}
\usepackage{subcaption}
\usepackage{setspace}
\usepackage[hidelinks]{hyperref}
\usepackage{overpic}
\usepackage{booktabs}
\usepackage{colortbl}
\usepackage{vcell} 
\usepackage[figuresright]{rotating}
\usepackage{epstopdf}

\graphicspath{ {} }
\providecommand{\keywords}[1]
{
  \small	
  \textbf{\textit{Keywords---}} #1
}

\title{Predicting Football Match Outcomes with eXplainable Machine Learning and the Kelly Index
}
\author[1]{Yiming Ren}
\author[1]{Teo Susnjak}
\affil[1]{School of Mathematical and Computational Sciences, Massey University, Auckland, New Zealand }
\begin{document}
\maketitle

% Here goes the abstract
\begin{abstract} % abstract

In this work, a machine learning approach is developed for predicting the outcomes of football matches. The novelty of this research lies in the utilisation of the Kelly Index to first classify matches into categories where each one denotes the different levels of predictive difficulty. Classification models using a wide suite of algorithms were developed for each category of matches in order to determine the efficacy of the approach. In conjunction to this, a set of previously unexplored features were engineering including Elo-based variables. 

The dataset originated from the Premier League match data covering the 2019-2021 seasons. The findings indicate that the process of decomposing the predictive problem into sub-tasks was effective and produced competitive results with prior works, while the ensemble-based methods were the most effective.

The paper also devised an investment strategy in order to evaluate its effectiveness by benchmarking against bookmaker odds. An approach was developed that minimises risk by combining the Kelly Index with the predefined confidence thresholds of the predictive models. The experiments found that the proposed strategy can return a profit when following a conservative approach that focuses primarily on easy-to-predict matches where the predictive models display a high confidence level.

\end{abstract}

%%%%%%%%%%%%%%%%%%%%%%%%%%%%%%%%%%%%%%%%%%%%%%
%%                                          %%
%% The keywords begin here                  %%
%%                                          %%
%% Put each keyword in separate \kwd{}.     %%
%%                                          %%
%%%%%%%%%%%%%%%%%%%%%%%%%%%%%%%%%%%%%%%%%%%%%%


\maketitle

\section{Introduction}\label{sec1}

Due to the worldwide appeal of football, the football industry has occupied an unassailable position in the sports business since its inception. The popularity of the sport is continuing to increase and increasing sports fans are becoming involved in the football industry. At the same time, betting on sporting events, including football, has become a growth industry. The number of legal sports betting organisers and participants worldwide is increasing every year. 

Meanwhile, the pre-match predictability of the sport is fairly low with some attribution placed on the sheer length of the matches as well as the number of players involved  \cite{hucaljuk2011predicting}. Others \cite{bunker2022application} have argued that the lower predictability of football can be explained by its low-scoring natures as well as the higher competitiveness relative to other sports and the fact that football can have more outcomes than merely wins and losses, but also draws, which contribute to the uncertainty of predicting the outcome of the matches\cite{wunderlich2018betting}. Other reasons have also been suggested such as power struggles within the leadership of football clubs\cite{huggins2018match}, match-fixing\cite{huggins2018match}, referees with low ethical standards\cite{carpenter2012match} and tacit understandings between teams on both sides of the game simultaneously reduce the predictability of football matches. The limited ability to counter match-fixing of local football leagues\cite{park2019should} and illegal internet gambling\cite{hill2010critical} are contributing components to the uncertainty of match outcomes.

The drive for increasing the predictability of football matches and profitability in gambling contexts,  has encouraged researchers to work on devising effective strategies. These approaches have taken the form of statistical\cite{baio2010bayesian}, machine learning\cite{baboota2019predictive}, \cite{fialho2019predicting}, natural language processing of football forums\cite{beal2021combining} as well as in-play image processing from camera footage of players\cite{lopes2014predicting} have been used in the attempt to  create more accurate predictive systems of football matches. However, even in recent years, research efforts have realised limited and incremental successes at predicting the outcome of pre-match football matches. 

In contrast to other sports, such as horse racing, there is no significant difference in the predictive accuracy between the opinions of the general public on internet forums, or the judgement of football experts and that of sophisticated machine learning methods\cite{brown2019wisdom} which underscores the challenge of the task. 
Even with access to rich sources of raw data and information, as well as the expert judgement of pundits,  the accuracy of the best predictions from literature after converting odds set by bookies into probabilities of events reaches  $\sim$55\%.

\subsubsection*{Contribution}

This study joins the growing body of literature that is attempting to improve the predictive accuracy of football matches using machine learning techniques, demonstrating the proposed methods on three seasons of Premier League matches in Europe sourced from publicly available datasets. In this work, we use a range of new features and algorithms that have not yet been used on this problem domain, and we employ tools from the explainable AI field in order to infuse the predictive models with interpretability and the ability to expose their reasoning behind the predictions. This study ultimately proposes a novel strategy for maximising the return on investment which relies on the Kelly Index in order to first categorise matches into levels of confidence with respect to their outcomes that are calculated on betting odds. We show how the decomposition of the problem with the aid of the Kelly Index and subsequent machine learning can indeed form an investment strategy that returns a profit.

\section{Related work}

This review divides previous studies in predicting the outcome of football matches into the target type of outcome (win/loss/draw) and those that have attempted at predicting the final scores of matches. However, because of the emergence of new betting options on betting platforms, predictions have emerged about whether or not both teams in the match will score goals\cite{da2022forecasting} \cite{inan2021using}. In fact, researchers now rarely predict the score of a match, as the uncertainty of the score and the existence of draws make score prediction  approach challenging and the prediction accuracy low, even with the inclusion of betting odds data\cite{reade2021evaluating}. The methods used for predicting the result of football matches fall into 3 categories. In the literature, statistical models, machine learning algorithms and rating systems have been explored.

\subsection*{Statistical and rating-system approaches}

\citet{inan2021using} fitted the team's offensive and defensive capabilities calculated from the goals made and conceded by teams each week to a Poisson distribution in his study while also taking  the home advantage into account. After fitting the team's offensive and defensive capabilities, home team advantage and betting parameters into a Poisson model, \citet{egidi2018combining} has developed a method of testing the prediction results to obtain more accurate predictions.

\citet{robberechts2018forecasting} used ELO ratings in predicting the result of FIFA World Cup matches. this approach considers the team's earlier result (whether it won or not) as the team's performance, and then gradually adjusts for subsequent results to obtain the final team performance rating. In order to identify and adjust a team's score, \citet{beal2021combining} used goals  by the team. They also considered the strength of the opponent of the match which is the source of the data to correct the team performance scores. In addition, \citet{constantinou2019dolores} used a hybrid Bayesian network to develop a dynamic rating system. The system corrects the team performance ratings by assigning greater weight to data on the results of the most recent matches that occurred. The researchers found that the study that uses scoring systems to predict the results of a football match usually uses databases with less content. These approaches need a small amount of data to extract features that can be used for prediction. 

\citet{koopman2019forecasting} used both a bivariate Poisson distribution model and a team performance rating system in their research. They held home team advantage constant when predicting National League outcomes but varied the team's offensive and defensive capabilities over time. The results proved that there was no major difference in prediction between the two methods and that both could make good predictions. However, the team performance rating system was ineffective at predicting draws. Most of the above research on the use of rating systems noted the near absence of ties in the predicted results. %But not all scoring systems are used directly to forecast the result of the match. 
\citet{berrar2019incorporating} used team performance scores as a predictive feature for machine learning.

\subsection*{Machine learning approaches}

Efforts to improve the predictability of football have focused on two aspects. They have either attempted to use more sophisticated machine learning algorithms, or on feature engineering in order to develop more descriptive variables. Sometimes both aspects have been pursued. Table \ref{Prediction accuracy of different studies} summarises all the key studies and their attributes.

\subsubsection*{Algorithms}

A mixture of more simplistic approaches like Linear Regression\cite{prasetio2016predictingLR}, K-nearest neighbors\cite{brooks2016usingKNN} and Decision Trees\cite{yildiz2020applyingDT} have been explored. Subsequently, the algorithms have increased in their sophistication with the use of Random Forests\cite{groll2019hybridRF}, Support Vector Machines\cite{oluwayomi2022evaluationSVM},  Artificial Neural Networks\cite{bunker2019machineANN} and boosting Boosting\cite{hubavcek2019learningBoosting}. More recently, Deep learning methods have emerged with the use of 
Convolutional Neural Networks \cite{hsu2021usingCNN} and LSTM \citet{zhang2021sports,malini2022deep} where the authors concluded that the LSTM model displayed superior characteristics to traditional machine learning algorithms and artificial neural networks in predicting the results of matches \cite{malini2022deep}.

\subsubsection*{Features}

Literature has shown that some of the effective features so far are half-time goal data as used by \citet{yekhande2582predicting}, first goal team data and individual technical behaviour data used by \citet{parim2021prediction}, ball possession and passing over data used by \citet{bilek2019predicting}, key player position data used by \citet{joseph2006predicting}. 

 Kınalıoğlu and Coskun\cite{kinaliouglu2020prediction} compared predictive models of six machine learning algorithms and evaluated models with different hyperparameters settings using a large number of model evaluation methods. \citet{beal2021combining} improved prediction accuracy by $\sim$7\% by extracting expert and personal opinions published on self-published media and combining environmental, player sentiment, competition, and external factors in their predictions.

Overall, studies using machine learning methods mention the need to use richer predictive features and validate them more robustly in future works. However, even newer studies have a tendency to reuse existing features and limited proposals have been made for innovating with novel features. 
Moreover, suggestions for new predictive features have primarily focused on ones that are not available in public datasets \cite{wunderlich2018betting} such as player transfer data, injuries, expert advice, and psychological data. 

Direct prediction of odds are rarely researched in this context, this is because the calculation of the odds itself contains the bookmaker's prediction of the results of the matches\cite{zeileis2018probabilistic}. Using odds to predict the outcome of a match requires a combination of machine learning or statistical methods. \citet{zeileis2018probabilistic} and \citet{wunderlich2018betting} both reverse the odds into the probability of a team winning and fit it into the ELO team performance scoring system to predict the outcome of the match. \citet{vstrumbelj2014determining} used traditional linear regression models and Shin models to predict the probability of winning directly from the odds, with the aim of inverting the bookmaker's method of calculating odds based on the betting.

\begin{sidewaystable}[]\tiny
\caption{\label{Prediction accuracy of different studies}Prediction accuracy of different studies}
\resizebox{\textwidth}{30mm}{
\begin{tabular}{lllllllll}
\hline
Study &
  Competition &
  Features &
  Non-pre-match features &
  Best Algorithm &
  Accuracy &
  Test set duration &
  Matches &
  Class \\ \hline
\citet{joseph2006predicting}\citeyear{joseph2006predicting} &
  Tottenham Hotspur Football Club &
  7 &
  None &
  Bayesian networks &
  58\% &
  1995-1997(2 years) &
  76 &
  3 \\
\citet{constantinou2019dolores} \citeyear{constantinou2019dolores} &
  52 football leagues &
  4 &
  None &
  Hybrid bayesian networks &
  - &
  2000-2017(17 years) &
  216,743 &
  3 \\
\citet{hubacek2019score}\citeyear{hubacek2019score} &
  52 football leagues &
  2 &
  None &
  Double Poisson &
  48.97\% &
  2000-2017(17 years) &
  218,916 &
  3 \\
\citet{mendes2020comparing}\citeyear{mendes2020comparing} &

  6 Leagues &
  28 &
  None &
  Bagging &
  51.31\% &
  2016-2019(3 years) &
  1,656 &
  3 \\
\citet{danisik2018football}\citeyear{danisik2018football} &

  5 Leagues &
  139 &
  None &
  LSTM regression &
  52.5\% &
  2011-2016 &
  1520 &
  3 \\
\citet{berrar2019incorporating}\citeyear{berrar2019incorporating} &

  52 football leagues &
  8 &
  None &
  KNN &
  53.88\% &
  2000-2017(17 years) &
  216,743 &
  3 \\
\citet{herbinet2018predicting}\citeyear{herbinet2018predicting} &

  5 leagues &
  6 &
  None &
  XGBoost &
  54\% &
  2014-2016(2 years) &
  3,800 &
  3 \\
\citet{carloni2021machine}\citeyear{carloni2021machine} &

  12 countries &
  47 &
  None &
  SVM &
  57\% &
  2008-2020(12 years) &
  49,319 &
  3 \\
\citet{baboota2019predictive}\citeyear{baboota2019predictive} &

  English Premier League &
  12 &
  None &
  Random Forest &
  57\% &
  2017-2018(1 year) &
  380 &
  3 \\
\citet{chen2019neural}\citeyear{chen2019neural} &

  La Liga &
  34 &
  None &
  Convolution neural network &
  57\% &
  2016-2017(1 year) &
  380 &
  3 \\
\citet{esme2018prediction}{}\citeyear{esme2018prediction} &

  Super League of Turkey &
  17 &
  None &
  KNN &
  57.52\% &
  2015-2016(half year) &
  153 &
  3 \\
\cite{rodrigues2022prediction}\citeyear{rodrigues2022prediction} &

  English Premier League &
  31 &
  None &
  SVM &
  61.32\% &
  2018-2019(1 years) &
  380 &
  3 \\
\citet{alfredo2019football}\citeyear{alfredo2019football} &

  English Premier League &
  14 &
  Half-time results &
  Random Forest &
  68,16\% &
  2007-2017(10 years) &
  3,800 &
  3 \\
\citet{pathak2016applications}\citeyear{pathak2016applications}&

  English Premier League &
  4 &
  None &
  Logistic Regression &
  69.5\% &
  2001-2015(14 years) &
  2280 &
  2 \\
\citet{prasetio2016predictingLR}\citeyear{prasetio2016predictingLR}&

  English Premier League &
  13 &
  Ball possession, Distance run &
  Logistic Regression &
  69.51\% &
  2015-2016(1 year) &
  380 &
  3 \\
\citet{danisik2018football}\citeyear{danisik2018football} &

  5 leagues &
  139 &
  None &
  LSTM regression &
  70.2\% &
  2011-2016 &
  1520 &
  2 \\
\citet{ievoli2021use}\citeyear{ievoli2021use} &

  UEFA Champions League &
  29 &
  Number of passes and other features &
  BLR &
  81\% &
  2016-2017(1 year) &
  75 &
  3 \\
\citet{azeman2021prediction}\citeyear{azeman2021prediction} &

  English Premier League &
  11 &
  Shooting, fouls and other 8 features &
  Multiclass decision forest &
  88\% &
  2005-2006(1 year) &
  380 &
  3 \\ \hline
\end{tabular}
}
\end{sidewaystable}

\subsection*{Summary of literature}

The reported accuracies by existing studies as summarised in Table~\ref{Prediction accuracy of different studies} are chiefly determined first by the nature of the features used, and specifically if these are generated in real-time during in-play match outcome predictions. Features generated in-play are more deterministic and thus result in considerably higher accuracies than more static pre-match features when predicting outcomes. Therefore these kinds of studies are different in nature than those that simply rely on pre-match prediction. 

Next, the accuracy is strongly affected by the nature of the predictive problem in the manner in which it is formulated. Models that attempt to predict a simple win or loss scenario are more accurate than those that also attempt to predict draws. This is expected since all predictive problems become more complex as the number of categories being predicted increases. For this reason, a number of researchers have opted for eliminating the prediction of tied matches.

\begin{figure}[h]
\centering
\includegraphics[width=0.9\textwidth]{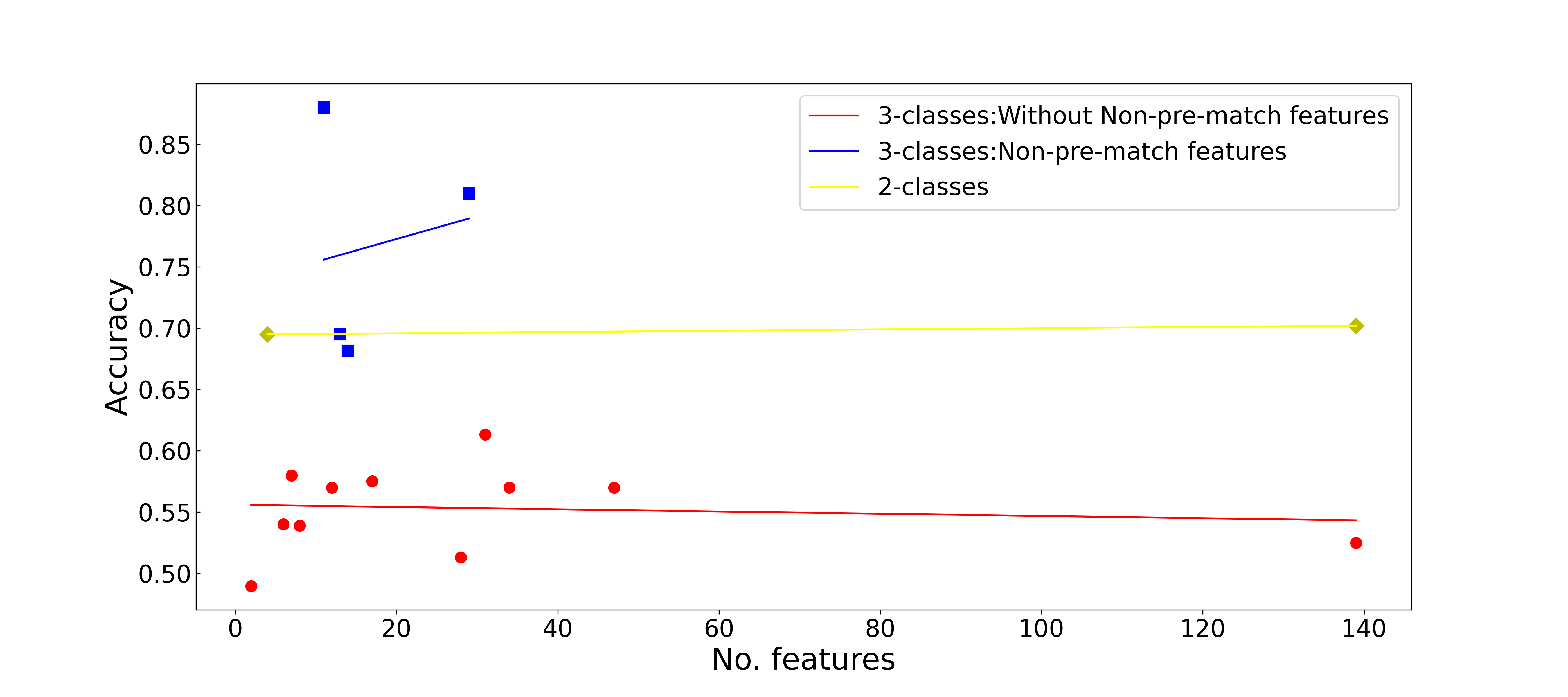}
\caption{The plot of the number of features against prediction accuracy for all types of literature}
\label{fig:Feature-Acc}
\end{figure}

In addition, the size of the datasets also affects the overall capability of the models, however, the number of features used in football match prediction has a small effect on prediction accuracy (Figure~\ref{fig:Feature-Acc}). The figure highlights that the accuracy of pre-match prediction is consistently below 60\%. However, when predicting pre-match results over multiple seasons, the prediction accuracy was almost always below 55\%. The researchers were unable to break the bottleneck in predicting football matches with more advanced algorithms and scoring systems.

When predictions are made over a large time span, for example, predicting the results of a game over 10 seasons, then these models will be less accurate than those predicting the results of a game over 1 season. 
Furthermore, the choice of machine learning method does not seem to have a significant impact on the accuracy of the model.

\subsubsection*{Research question}

In light of the literature, this study poses the following research questions:

\begin{enumerate}
    \item Is it possible to use public data to find out which matches are difficult to predict and which results are already certain before the match?
    \item Can prediction accuracy be improved by training models using hard-to-predict matches or easy-to-predict matches?
    \item Which machine learning predictions perform best for hard-to-predict matches and easy-to-predict matches?
    \item Is it possible to increase returns by applying different weights to hard-to-predict matches and easy-to-predict matches when investing?
\end{enumerate}

\section{Methodology}

\subsection{Machine learning algorithms}
The machine learning methods used  are drawn from open-source machine learning libraries. These include Scikit-learn\cite{pedregosa2011scikit}, XGboost and CatBoost.

\paragraph*{Logistic regression} Logistic regression predicts the probability on the basis of linear regression. It converts predictions to probabilities and then implements classification.
The algorithm demonstrates a strong ability to deal with data with a strong linear relationship between features and labels. Logistic regression also has the advantage of being noise-resistant and performs well on small data sets. \citet{prasetio2016predictingLR} used logistic regression in predicting the English Premier League. Using in-play features, the authors obtained a prediction accuracy of 69\%.

\paragraph*{Decision Tree}  This algorithm builds a Decision Tree \cite{quinlan1986induction} by choosing features that have the greatest information gain. The advantage is that it does not require any domain knowledge or parameter assumptions and is suitable for high-dimensional data. However, this algorithm is prone to overfitting and tends to ignore correlations between attributes. The use of a small number of features in the Decision Tree methods can prevent overfitting when predicting the results of football matches. The Decision Tree approach may therefore produce a more explanatory and thus informative model.

\paragraph*{Random Forest} This is an ensemble-based method  which combines  multiple Decision Trees. The Random Forest approach integrates multiple Decision Trees and builds each one using a non-backtracking approach. Each of these Decision Trees relies on independent sampling and has the same randomness in data selection as the other trees. When classifying the data, each tree is polled to return the category with the most votes, giving a score for the importance of each variable and assessing the importance of each variable in the classification. Random Forests tend to perform better than Decision Trees in preventing overfitting.

The correlation between the model features and the test predictions in the Random Forest approach is not strong. For poorly predicted football match results, this may be an advantage of the Random Forest approach. Using a training set containing a large number of features can result in a generated model with strong predictive power on a small number of them and weak predictive power on others. The classification criteria are different for each Decision Tree in the Random Forest model. This can also be a good way to reduce the correlation between Decision Trees.

\paragraph*{k-Nearest Neighbor} The k-Nearest Neighbor(KNN) method classifies samples by comparing a particular sample with its \textit{k} closest data points in a given dataset and therefore does not generate a classifier or a model  \cite{peterson2009k}.

KNN methods have an excellent performance when using data from in-play contexts to predict the results of contests\cite{esme2018prediction}. However, the method needs to consider all the data in the sample when generating the prediction model. Such an approach creates a strong risk of overfitting in the prediction of pre-match results in competitive sports. Nevertheless, it is still not known how well the KNN method can perform for hard-to-predict matches.

\paragraph*{Gradient Boosting} The idea behind gradient boosting is to iteratively generate multiple weak models and then add up the predictions of each weak model. Each residual calculation increases the weight of the wrongly split samples, while the weights of the split pairs converge to zero, so the generalisation performance is better. The importance of the features of the trained model can be extracted. 

In previous studies\cite{alfredo2019football} of match prediction gradient boost methods have often failed to produce the best models. However, the predictive power of the method in hard-to-predict matches cannot be denied outright.

\paragraph*{XGBoost} XGBoost extends and improves upon the Gradient Boosting. The XGBoost algorithm is faster. It takes advantage of the multi-threading of the CPU based on traditional Boosting and introduces regularisation to control the complexity of the model. Prior to the iteration, the features are pre-ranked for the nodes and by traversing them the best segmentation points are selected, which results in lower complexity of data segmentation.

The XGBoost method is popular amongst machine learning researchers. However, the method does not perform well in capturing high-dimensional data, such as images, audio, text, etc. For predicting football matches with based on lower-dimensional data, XGBoost may produce promising models.

\paragraph*{CatBoost} 
CatBoost algorithm is a type of gradient boosting algorithm. In contrast to the gradient boosting algorithm, this algorithm can also achieve excellent results using data that has not been feature engineered. It does better than gradient boosting in preventing overfitting. The CatBoost method has not been explored in previous studies of football match result prediction.

\paragraph*{Voting and Stacking} The voting classifier can vote on the results of different models, with the majority deciding the outcome. The Voting classifier is divided into Hard and Soft implementation. The Hard voting is votes on the results obtained using multiple machine learning methods, with the majority getting the result. Soft voting is the process of calculating the mean of the classification probabilities of the different methods, and finally selecting the one with the highest mean as the prediction result.

The stacking algorithm refers to a hybrid estimator approach in which a number of estimators are fitted separately on training data, while the final estimator is trained using the stacked predictions of these basic estimators. However, the stacking method uses cross-validation which is not favoured in the field of football result prediction when training the model.

Voting and stacking methods have produced the best prediction models in other studies of athletic match result prediction\cite{li2019high}. However, in football these two methods are rarely used. This study uses both methods in an attempt to find the most suitable model for prediction in difficult-to-predict matches.

\subsection{Dataset characteristics and feature engineering}

The Premier League data used in this study was sourced from the website https://www. footballdata.co.uk/englandm.php. The results from the 2001 to 2018 seasons were used to develop features based on ELO ratings, attack and defence ratings and home-team and away-team winning percentages. Data from the 2019 to 2021 seasons were used for testing. The raw data was recorded using a total of 107 columns of data for the 2019 to 2021 seasons.

The Premier League system divides the season into 38 rounds and guarantees that each of the 20 teams will play once in each of the 10 matches in each round. After 380 matches per season, each team plays 19 times in the home position and 19 times in the away position. This study predicts the results of 3 seasons, implying a total of 1140 predicted matches. 

The Premier League is based on a points system, with the winning team gaining three points after the match and the two drawing teams gaining one point each. 24 teams are entered for the 2019-2021 season, all of which appeared in the 2001-2018 season. However, due to differences in the raw data on each season on bookmaker odds records, in fact, only 21 of the 83 columns of data were used in the 2019-2021 season predictions. The records are for the six European bookmakers and the European average odds respectively.

The premise of this study for feature engineering is the assumption that all matches occur continuously. 

A total of 52 features were generated and explored in this project. Their names and descriptions are given in Table~\ref{Table 1. Description of features}.

\begin{table}[]
\centering
\caption{\label{Table 1. Description of features}Description of features}
\fontsize{7pt}{9pt} 
\selectfont
\begin{tabular}{@{}ll@{}}
\toprule
Feature Name  & Description                                                                         \\ \midrule
AvgGoalDiff   & Average goal difference between the two teams in the previous six games             \\
TotalGoalDiff & Goal difference between the two teams in the previous six games                     \\
HomeELO       & ELO ratings for the home team                                                       \\
AwayELO       & ELO ratings for away teams                                                          \\
ELOsta        & Standard deviation of the ELO ratings of the two teams                              \\
ELOHomeW      & Probability of home team winning converted by ELO ratings                           \\
ELOAwayW      & Probability of away team winning converted by ELO ratings                           \\
ELODraw       & Probability of a draw occurring converted by ELO ratings                            \\
one\_ELO      & Probability of conversion from ELO ratings after PCA dimensionality reduction       \\
HomeHELO      & ELO ratings for the home team's half-time result                                    \\
AwayHELO      & ELO ratings for the away team's half-time result                                    \\
HELOSta       & Standard deviation of the half-time ELO ratings of the two teams                    \\
ELOHHomeW     & Probability of home team winning converted by half-time ELO ratings                 \\
ELOHAwayW     & Probability of away team winning converted by half-time ELO ratings                 \\
ELOHDrawW     & Probability of a draw occurring converted by half-time ELO ratings                  \\
one\_HELO    & Probability of conversion from half-time ELO ratings after PCA dimensionality reduction                                                    \\
HomeTeamPoint & Current home team points in the season                                              \\
AwayTeamPoint & Current away team points in the season                                              \\
PointDiff     & Current point difference between home and away teams in the season                  \\
AvgHOddPro    & Average of pre-match home team odds offered by all bookmakers in the available data \\
AvgAOddPro    & Average of pre-match away team odds offered by all bookmakers in the available data \\
AvgDOddPro    & Average of pre-match draw odds offered by all bookmakers in the available data      \\
one\_Odd\_Pro & Average odds after PCA dimensionality reduction                                     \\
HomeOff       & Rating of the home team's offensive capabilities                                    \\
HomeDef       & Rating of the home team's defensive capabilities                                    \\
AwayOff       & Rating of the away team's offensive capabilities                                    \\
AwayDef       & Rating of the away team's defensive capabilities                                    \\
Offsta        & Difference in offensive capability rating                                           \\
Defsta        & Difference in defensive capability rating                                           \\
AvgShotSta   &  Standard deviation of shots on goal for both sides in six matches          \\
AvgTargetSta & Standard deviation of shots on target for both sides in six matches \\
ShotAccSta    & Standard deviation of shot accuracy between the home team and the away team         \\
AvgCornerSta & Standard deviation of corners for both sides in six matches         \\
AvgFoulSta   & Standard deviation of fouls for both sides in six matches          \\
HomeHWin      & Home team's winning percentage in home games                                        \\
HomeHDraw     & Home team's draw percentage in home games                                           \\
AwayAWin      & Away team's winning percentage in away games                                        \\
AwayADraw     & Away team's draw percentage in away games                                           \\
HomeWin       & Home team's win percentage in all previous matches                                  \\
HomeDraw      & Home team's draw percentage in all previous matches                                 \\
AwayWin       & Away team's win percentage in all previous matches                                  \\
AwayDraw      & Away team's draw percentage in all previous matches                                 \\
LSHW          & Home team win percentage last season                                                \\
LSHD          & Home team draw percentage last season                                               \\
LSAW          & Away team win percentage last season                                                \\
LSAD          & Away team draw percentage last season                                               \\
Ysta          & Difference in the number of yellow cards between the home team and the away team    \\
Rsta          & Difference in the number of red cards between the home team and the away team       \\
StreakH       & Home team winning streak index                                                      \\
StreakA       & Away team winning streak index                                                      \\
WStreakH      & Home team weighted winning streak index                                             \\
WStreakA      & Away team weighted winning streak index  
                                        \\
\midrule                            
\end{tabular}
\end{table}

\subsubsection*{ELO Rating}

The ELO rating was first introduced by ELO\cite{ELO1978} to assess the ability of players in chess competitions and has been extended to the assessment of the performance of players or teams in many sports. The rationale is to achieve a dynamic evaluation of a particular player's ability by comparing the ability of the player in a recent match with that player's past performance. The time-series nature of football matches makes the  ELO model suitable for scoring the performance of football teams\cite{jain2021sports}.

The score of the participating teams in a football match is determined by their performance in past matches (Eq~\ref{Eq.1}) and the rating of their performance in the current match of the competition (Eq`\ref{Eq.2}) are as follows:

\begin{align}\label{Eq.1}
    E_{}^{H} & = \cfrac{1}{1 + \cfrac{c^{R^{H} -R^{A}} }{d}}  \quad\quad\quad E_{}^{A}  = 1-E_{}^{H}
\end{align}
    
\begin{align}\label{Eq.2}
S_{}^{H} & = \left\{\begin{matrix} 
  1 \qquad  Home\quad team\quad win\\
  0.5 \qquad  Draw\\
  0\qquad  Away\quad team\quad win
\end{matrix}\right. \qquad 
\end{align}

\begin{align}\label{Eq.3}
S_{}^{A} = 1 -S_{}^{H}
\end{align}

where $E^H$ and $E^A$ are the performance ratings that teams are supposed to have in that match, $S^H$ and $S^A$ are the actual performance ratings of two teams. $R^H$ and $R^A$ are the ratings that the two teams have constantly revised to represent their own team's ability, and $c$ and $d$are two constants based on the scale of scoring. In this study $c$ was set to 10 and $d$ to 400 to ensure that the effect of the new matches' results on the previous ELO ratings was kept in a balanced range.. The scoring of the two teams is appropriately corrected by the error between the ability of the two teams during the newly occurring match and previously assessed performance, using the following:

\begin{align}
R^{'H} = R^{H} +k(S^{H}-S^{E})
\end{align}

where $k$ indicates the magnitude of the correction. $k$ is used to determine the effect of the results of the new competition on the scoring. $k$ is calculated as follows:

\begin{align}
k = k_0(1+\delta)^{\gamma }
\end{align}

where $\delta$ is the absolute goal difference and $k_0$ and $\gamma$ are set to 10 and 1.

The performance ability of teams was first assessed in this study using match results from the 2001-2005 English Premier League season. Teams were ranked and assigned ratings by comparing the total scores and wins of teams in the 2001-2002 season. The team ratings were corrected using match data from the 2002-2017 season. 2018-2019 season was formally characterised using the ratings generated by the ELO model, and each team's rating was subsequently corrected for the previous match results.

In a subsequent study on ELO ratings, a regression model was generated to predict wins and losses by the ELO rating profile of the teams from both sides before the match\cite{hvattum2010using}. The model has a high degree of confidence among those involved in football gambling. The probability of different match outcomes occurring can be simply predicted using the ELO scores of the teams on both sides of the match using the following formula:

\begin{align}
P_{H} & = 0.448+(0.0053*(E^{H} -E^{A} )) 
\end{align}
\begin{align}
P_{A} & = 0.245+(0.0039*(E^{A} -E^{H} )) 
\end{align}
\begin{align}
P_{D} = 1- (P_{H}+P_{A})
\end{align}

where $P^H$ indicates the probability of the home team victory, $P^A$ indicates the probability of the away team victory occurring and $P^D$ indicates the probability of a draw situation occurring.

\subsubsection*{Offensive and defensive capabilities}

The Offensive and Defensive Model (ODM) \cite{govan2009offense} is used to generate a team's offensive and defensive abilities. The calculation formula (Eq~\ref{Eq.9}) for scoring given in the introduction of the ODM model by \citet{govan2009offense} is as follows:

\begin{align}\label{Eq.9}
o_j = \sum_{i = 1}^{n} \frac{{A_i}_j}{d_i}  \qquad dj = \sum_{i = 1}^{n} \frac{{A_j}_i}{o_i}
\end{align}

where ${A_i}_j$ is a goal scored by team $j$ in a match between teams from both sides of the match, $o_j$ is the offensive rating and $d_j$ is the defence rating. The equation shows that the two ratings affect each other.  We directly assess the offensive and defensive ratings of a team one season and apply them to the next season. A team with high goal-scoring has a high offensive rating. Likewise, a team with a high number of goals conceded has an inefficient defensive rating. 

For teams new to the season, we extrapolate their offensive and defensive ratings based on that team's ranking in goals scored and goals against during the season. This is calculated as follows:

\begin{align}
o_j = \frac{XG_2+X_2G_1-X_1G_1-X_1G_2}{X_2-X_1} 
\end{align}

where $O_j$ represents the new team's offensive score. $G$ is the goals scored by the new team during the season. $G1$ and $R1$ are the goals scored and offensive rankings of the teams that are one place above the new team in the season goal rankings. $G2$ and $R2$ are the goals scored and rankings of the teams that are one place below. The defensive rating is calculated in the same way.

\subsubsection*{Streak Index}

The Streak Index represents a team's capacity to go on a winning streak which captures a team's most recent form. The index was generated by \citet{baboota2019predictive}  in predicting the results of the Premier League for the two seasons 2014-2016. the Streak Index is further divided into two categories, the first of which directly measures the team's form over the previous $k$ games, with the following formula:

\begin{align}
S = (\sum_{p=j-k}^{j-1} res_{p} )/3k
\end{align}

where $j$ indicates the number of matches to be predicted and $res_p$ indicates the team's score in a particular match, where a win and a draw correspond to 3 points and 1 point respectively. No points are awarded for losses. In the second category, the Streak Index, is weighted and normalised based on the first category. Relatively low weights are given to matches that occur further back in time. The formula for this is as follows:
\begin{align}
\omega _{S} = \sum_{p=j-k}^{j-1} 2\frac{p-(j-k-1)res_{p} }{3k(k-1)} 
\end{align}

The Streak Index is logically problematic when calculating across seasons. This means that when calculating the Streak Index for the first few games of a season, the data used is from the last season's end. The changes in team form that occur during the intervening season can compromise the effectiveness of the Streak Index.

\subsubsection*{Principal Component Analysis}

As the outcome of a football match includes a draw, bookmakers also offer odds in the event of a draw. This requires a feature reduction method to turn three-dimensional data into one-dimensional data.

This study uses the Principal Component Analysis (PCA) \cite{wold1987principal}  to reduce the dimensionality of the data. This function linearly reduces the dimensionality on the original data by projecting the original data to an alternative dimension where the data loss is minimised. 
PCA is generally used to process and compress high-dimensional data sets. The main objective is to reduce the dimensionality of a large feature set into one that is most essential with the tradeoff that the new features lose interpretability. 

\subsubsection*{Kelly Index}

The Kelly index\cite{thorp1975portfolio} was used as a tool to classify the matches in  the dataset into categories that define different levels of predictability. 
In literature, the Kelly Index was originally used to calculate the flow-through rate of electronic bits\cite{kelly1996synchronization}. However, due to its probabilistic nature and its similarity to the nature of betting, the Kelly Index is also widely used by bettors.

Each bookmaker calculates the Kelly Index before a match and makes it available to members who gamble. 
The Kelly Index can be calculated from the available data using the following formula:

\begin{align}
K_{H}=(O_{H}/avgO_{H} )F_{99}   &\\K_{A}= (O_{A}/avgO_{A} ) F_{99} &\\K_{D}=(O_{D}/avgO_{D} )F_{99}
\end{align}

where $K_{H}$, $K_{A}$, and $K_{D}$ indicate the three results of matches in terms of the Kelly Index. $avgO_{H}$, $avgO_{A}$, and $avgO_{D}$ indicate in our dataset the average European betting market odds for each of the three results of the matches. $F_{99}$ indicates the return rate of the European average odds, which is calculated by the following formula:

\begin{align}
F_{99} =\frac{1}{\frac{1}{avgO_{H}}+ \frac{1}{avgO_{A}}+\frac{1}{avgO_{D}}} 
\end{align}

If a bookmaker has set lower odds on an event occurring than any other bookmaker,  this means that the bookmaker is confident enough to believe that a particular outcome will occur. This confidence may come from the bookmaker's perception of possessing more effective prediction models, having access to more extensive sources of information, or illegal black market trading. The bookmaker as a dealer is willing to set the lowest odds on a more likely event to maximise profits. Therefore a bookmaker's Kelly Index is the most likely event to occur. At the same time, if there is a low Kelly Index at one bookmaker, there will be a high Kelly Index for other events accordingly. However, there is no guaranteed correlation between the level of the Kelly Index and the result of a match as this is just a technique for understanding what odds have been determined to be in the best interest of the bookmaker for maximising their returns.

\begin{figure}[h]
\centering
\includegraphics[width=0.9\textwidth]{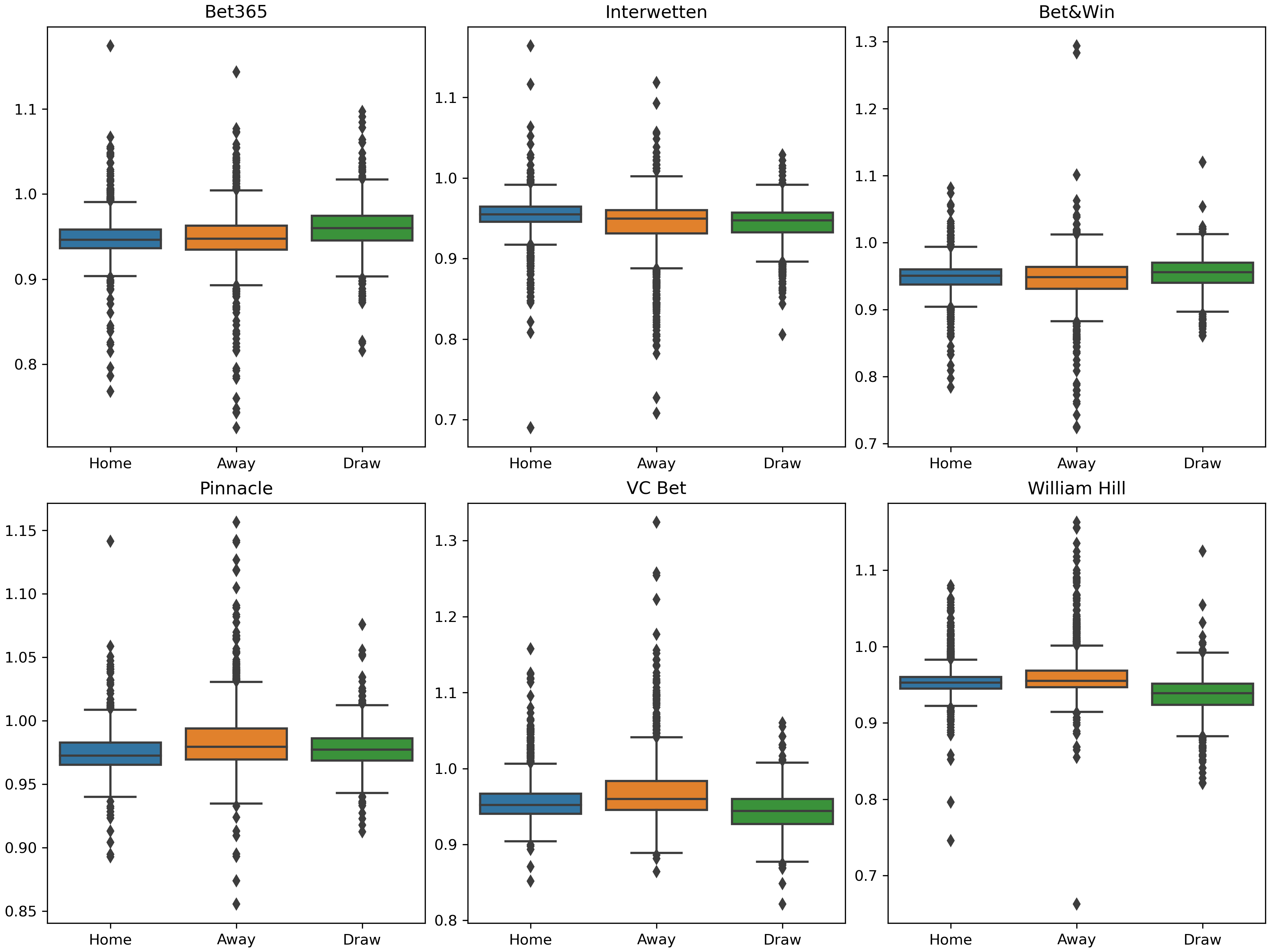}
\caption{The distribution of the Kelly Index of the six European bookmakers for the 2019-2021 season.}
\label{fig:The distribution of the Kelly Index of the six European bookmakers for the 2019-2021 season}
\end{figure}

As shown in Figure ~\ref{fig:The distribution of the Kelly Index of the six European bookmakers for the 2019-2021 season}, the Kelly Indexes of these six bookmakers are mainly concentrated in the range of 0.9-1.0. In the setup of the predictive problem, the matches for the 2005-2021 season were divided into 3 categories with each one reflecting the confidence levels of their predictability. These were, matches with Kelly indexes greater than 1 (Type 1), matches with only one Kelly index greater than 1 (Type 2), and matches with no Kelly Index greater than 1 (Type 3). 

\subsection{Model Optimisation}

\subsubsection*{Hyperparameter tuning}

This study used the RandomizedSearchCV method from the scikit-learn library to select the most effective parameter values. All the parameter tuning values for the machine learning algorithms used in this study are shown in Table~\ref{Values for parameter tuning}.

\begin{table}[hbt]
\centering
\caption{\label{Values for parameter tuning}
Values for parameter tuning
}
\fontsize{7pt}{9pt} 
\selectfont
\begin{tabular}{@{}lll@{}}
\midrule
Algorithm                                                       & Parameter                  & Range of values                           \\
\midrule
\vcell{CatBoost}                                                & \vcell{learning\_rate}     & \vcell{0.01,0.02,0.03,0.04}               \\[-\rowheight]
\printcelltop                                                   & \printcellmiddle           & \printcellmiddle                          \\
                                                                & iterations                 & 10,20,30,40,50,60,70,80,90                \\
                                                                & depth                      & 4,5,6,7,8,9,10                            \\
\vcell{Decision Tree}                                           & \vcell{criterion}          & \vcell{'gini', 'entropy'}                 \\[-\rowheight]
\printcelltop                                                   & \printcellmiddle           & \printcellmiddle                          \\
                                                                & max\_depth                 & 2,4,6,8,10,12                             \\
\vcell{Gradient boost}                                          & \vcell{min\_samples\_leaf} & \vcell{2,5,8}                             \\[-\rowheight]
\printcelltop                                                   & \printcellmiddle           & \printcellmiddle                          \\
                                                                & min\_samples\_split        & 3,5,7,9                                   \\
                                                                & max\_features              & Based on number of filtered features  \\
                                                                & max\_depth                 & 2,5,7,10                                  \\
                                                                & learning\_rate             & 0.1,1.0,2.0                               \\
                                                                & subsample                  & 0.5,0.8,1                                 \\
\textcolor[rgb]{0.125,0.129,0.141}{k-Nearest Neighbor}          & n\_neighbors               & Based on sample size                      \\
\vcell{\textcolor[rgb]{0.071,0.071,0.071}{Logistic regression}} & \vcell{solver}             & \vcell{liblinear}                         \\[-\rowheight]
\printcelltop                                                   & \printcellmiddle           & \printcellmiddle                          \\
                                                                & penalty                    & 'l1','l2'                                 \\
                                                                & class\_weight              & 1:2:1,3:3:4,4:3:3                         \\
\vcell{Random Forest}                                           & \vcell{min\_samples\_leaf} & \vcell{2,5,8}                             \\[-\rowheight]
\printcelltop                                                   & \printcellmiddle           & \printcellmiddle                          \\
                                                                & max\_features              & Based on number of filtered features  \\
                                                                & max\_depth                 & 2,5,7,10                                  \\
Stacking                                                        & Same as above algorithm    & Same as above algorithm                   \\
Voting                                                          & Same as above algorithm    & Same as above algorithm               \\
\hline
\end{tabular}
\end{table}

\subsubsection*{Feature selection}
Models developed with feature selection tend to have greater explanatory power, run faster, and have a decreased risk of over-fitting. In some cases, feature selection will improve the predictive power of models. With the exception of the two ensemble algorithms Voting classifier and Stacking, feature selection was used to enhance the models.

Feature selection decisions were made using Shapley Additive exPlanations\cite{lundberg2017unified} (SHAP). SHAP is based on the Shapley value\cite{winter2002shapley}, a concept derived from cooperative games, and is a common metric for quantitatively assessing the marginal contribution of users in an equitable manner. SHAP has been implemented as a visualisation tool for model interpretation that is able to score every feature in terms of its effects on the final outcome. Efficient feature selection can be achieved by removing features that are indicated to have low impacts on predictions. For multiclassification gradient boosting models where the SHAP explainer is not available, the Recursive Feature Elimination method in the scikit-learn library was used to remove the least important features recursively selected in the dataset.

In the feature selection process, after generating a model that has been modelled using all features and tuned with hyperparameters, the features in the validation set with an average SHAP value of less than zero were removed and the dataset was remodelled after the low-impact features were. This loop continues until the number of features is less than 2 or no features with a SHAP value less than zero appear. After the iterative process is completed, the model with the highest prediction accuracy in the validation set is selected as the prediction model for the test set.

\subsection{Model training and testing framework}

\subsubsection{Modelling process}

\begin{figure}[h]
\centering
\includegraphics[width=0.9\textwidth]{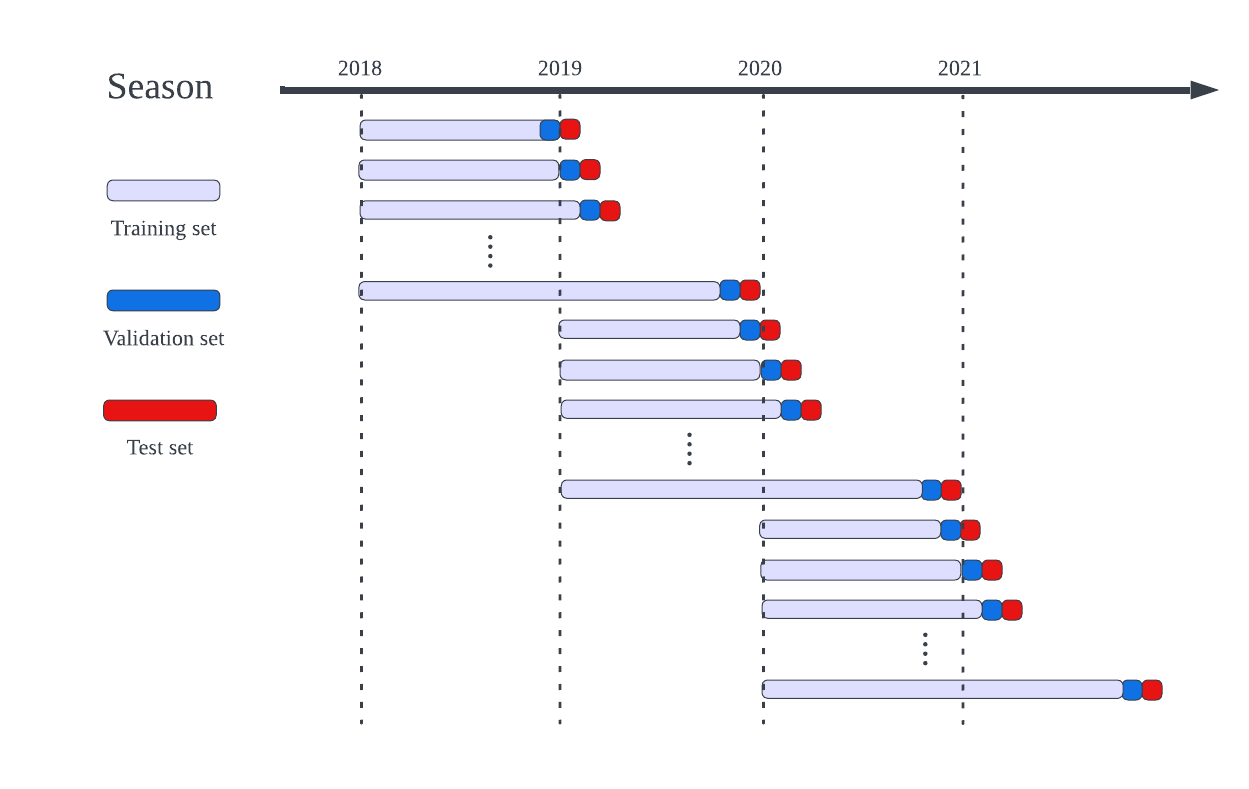}
\caption{Window extension strategies for predictive models}
\label{fig:Window extension strategies for predictive models}
\end{figure}

\begin{figure}[h]
\centering
\includegraphics[width=0.5\textwidth]{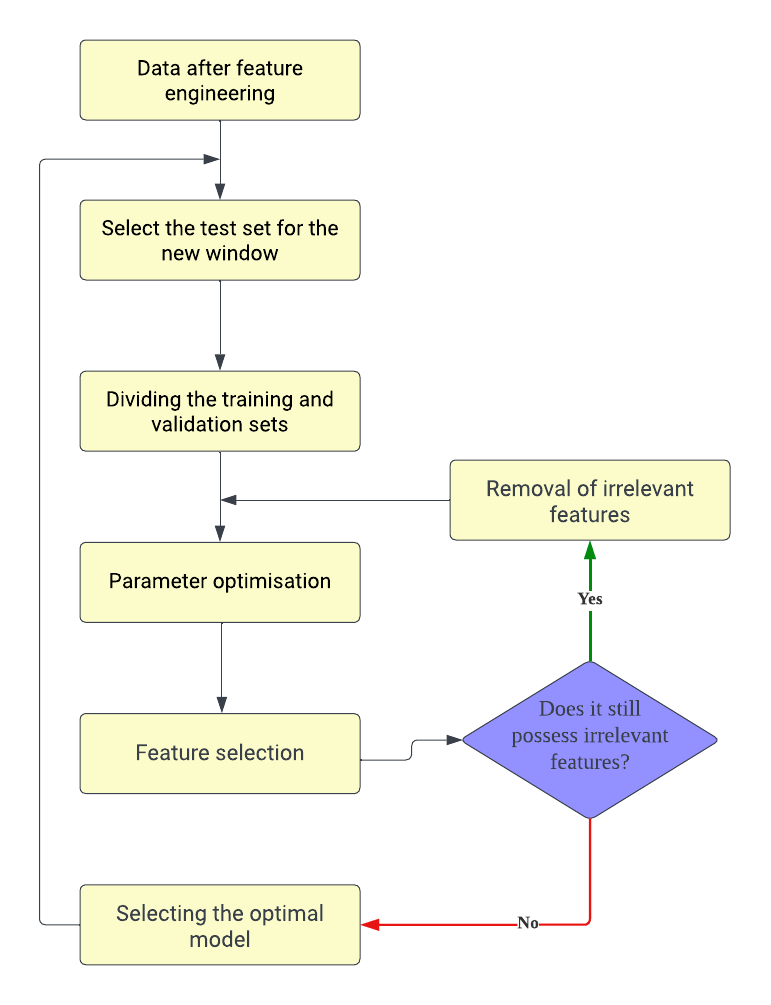}
\caption{Modelling process}
\label{fig:modelling process}
\end{figure}

This study uses an extended window approach to test the models in order to avoid the mistake of using information from future matches to predict past matches which is in contrast to a number of other studies on sports outcome prediction. \citet{bunker2019machineANN} outline in detail why  standard cross-validation is not suitable for evaluation of prediction models in this domain. The pattern of the training and testing sequences comprising windows of division is shown in Figure  ~\ref{fig:Window extension strategies for predictive models} while the modelling process is shown in Figure ~\ref{fig:modelling process}. 

Training sets from the immediately preceding season are used to make predictions for the next season. This strategy was followed since it has been shown that it is more informative to choose the most recent matches for predictions of upcoming matches. The reason for this is that the game of football, being a high-intensity competitive sport, advances over time in terms of its macro tactics and the detail of the players' actions\cite{smith2007nature}. As the number of different types of matches varies from season to season, the size of the test set is set at 20 to 30 matches depending on the number of matches of a particular type in the current season.

The validation set is used in order to select the best model during parameter tuning. When using models where parameter tuning has little impact on model quality (e.g. Voting Classifier) the validation set is not used and the part of the model that would otherwise be part of the validation set is merged with  the training set. The validation set is usually set to the same size as the test set for that window.

\subsubsection{Model evaluation}

This study evaluates the predictive ability of each classifier by using four metrics. These four reported metrics are accuracy, precision, recall and F1 Score. 
Accuracy is used as a primary metric for ranking the performances of the algorithms which the remaining metrics are used as a reference for completion.
These evaluation metrics are calculated as follows:

\begin{align}
Accuracy = \frac{TP+TN}{TP+FN+FP+TN}
\end{align}

where the denominator part indicates the number of all samples and the numerator part indicates the number of all correctly predicted samples.

\begin{align}
Precision = \frac{TP}{TP+FP}
\end{align}

where the denominator part indicates the number of samples for a particular result and the numerator part indicates the number of correctly predicted samples for that result species. The precision of the three classification questions is weighted by the proportion of each result in the sample using the precision of the three results.

\begin{align}
Recall = \frac{TP}{TP+FN} 
\end{align}

Theoretically, the three classification problems have the same recall and accuracy on a given result. The recall in this study uses macro-Recall without considering the weights of each result.

\begin{align}
F_{1} = \frac{2\times Precision \times Recall}{Precison + Recall} 
\end{align}

To evaluate the relative capability of the models, two baseline models were added for comparison. Both models are referenced from the Dummy Classifier class in the scikit-learn library. Both models ignore the data in the training set and make random predictions on the test set. Baseline model 1 is a random selection of one result from three results of matches as the predicted result. Baseline model 2 has a similar prediction process to baseline model 1, but predictions follow the distribution of each result in the test set.

\section{Result and discussion}\label{sec2}

\subsection{Match classification}\label{fm}

This study classified 1,140 matches in the English Premier League for the three seasons of 2019-2021 into three possible categories, namely win, loss, draw. Matches with multiple bookmakers having a Kelly Index greater than 1 were categorised as Type 1. Matches with only one bookmaker having a Kelly Index greater than 1 were categorised as Type 2. Matches with no bookmaker having a Kelly Index greater than 1 were categorised as Type 3. Their precise numbers in each season are shown in Table~\ref{Number of matches divided into different types}. The number of Type 1 matches decreases yearly, while the number of Type 2 matches increases yearly. This may indicate that bookmakers are generally becoming progressively more conservative in setting their odds.

\begin{table}
\centering
\caption{\label{Number of matches divided into different types}Number of matches according to different types and categories}
\label{Number of matches divided into different types}
\fontsize{7pt}{9pt} 
\selectfont
\begin{tabular}{lcccccc} 
\hline
\multirow{2}{*}{Category}  & \multirow{2}{*}{Season} & \multicolumn{4}{c}{Result Number}            & \multirow{2}{*}{Draw proportion}  \\
                           &                         & Home & Away & Draw & \multicolumn{1}{l}{All} &                                   \\ 
\hline
\multicolumn{1}{c}{Type 1} & 2019                    & 61   & 28   & 17   & 106                     & 16.0\%                           \\
                           & 2020                    & 50   & 43   & 18   & 111                     & 16.2\%                           \\
                           & 2021                    & 42   & 18   & 16   & 76                      & 21.1\%                           \\
\multicolumn{1}{c}{Type 2} & 2019                    & 35   & 28   & 18   & 81                      & 22.2\%                           \\
                           & 2020                    & 36   & 29   & 22   & 87                      & 25.3\%                           \\
                           & 2021                    & 55   & 42   & 27   & 124                     & 21.8\%                           \\
\multicolumn{1}{c}{Type 3} & 2019                    & 76   & 60   & 57   & 193                     & 29.5\%                           \\
                           & 2020                    & 58   & 81   & 43   & 182                     & 23.6\%                           \\
                           & 2021                    & 66   & 69   & 45   & 180                     & 25.0\%                           \\
\multicolumn{1}{c}{All}    & 2019                    & 172  & 116  & 92   & 380                     & 24.2\%                           \\
                           & 2020                    & 144  & 153  & 83   & 380                     & 21.9\%                           \\
                           & 2021                    & 163  & 129  & 88   & 380                     & 23.2\%       
                           \\
                           \hline
\end{tabular}
\end{table}

Nearly half of all 1140 matches belonged to the Type 3. In these matches, the bookmakers were not confident in setting odds that constituted an increased liability for them.
Ultimately, Type 3 matches produced the highest percentage of draws, while Type 1 matches had the lowest percentage of draws confirming the efficacy of the Kelly Index to reduce uncertainty to some degree. Although draws are difficult to predict, they do not account for more than a third of matches in any category. In addition, home-team wins in Type 1 matches were much higher than the number of away-team wins. However, away-team wins were higher in the Type 3 matches. Bookmakers consider both draws and home team advantages when formulating their predictions.  Bookmakers are more likely to reduce the odds of a home win when considering setting odds or home teams.

\begin{figure}[h]
\centering
\includegraphics[width=1.0\textwidth]{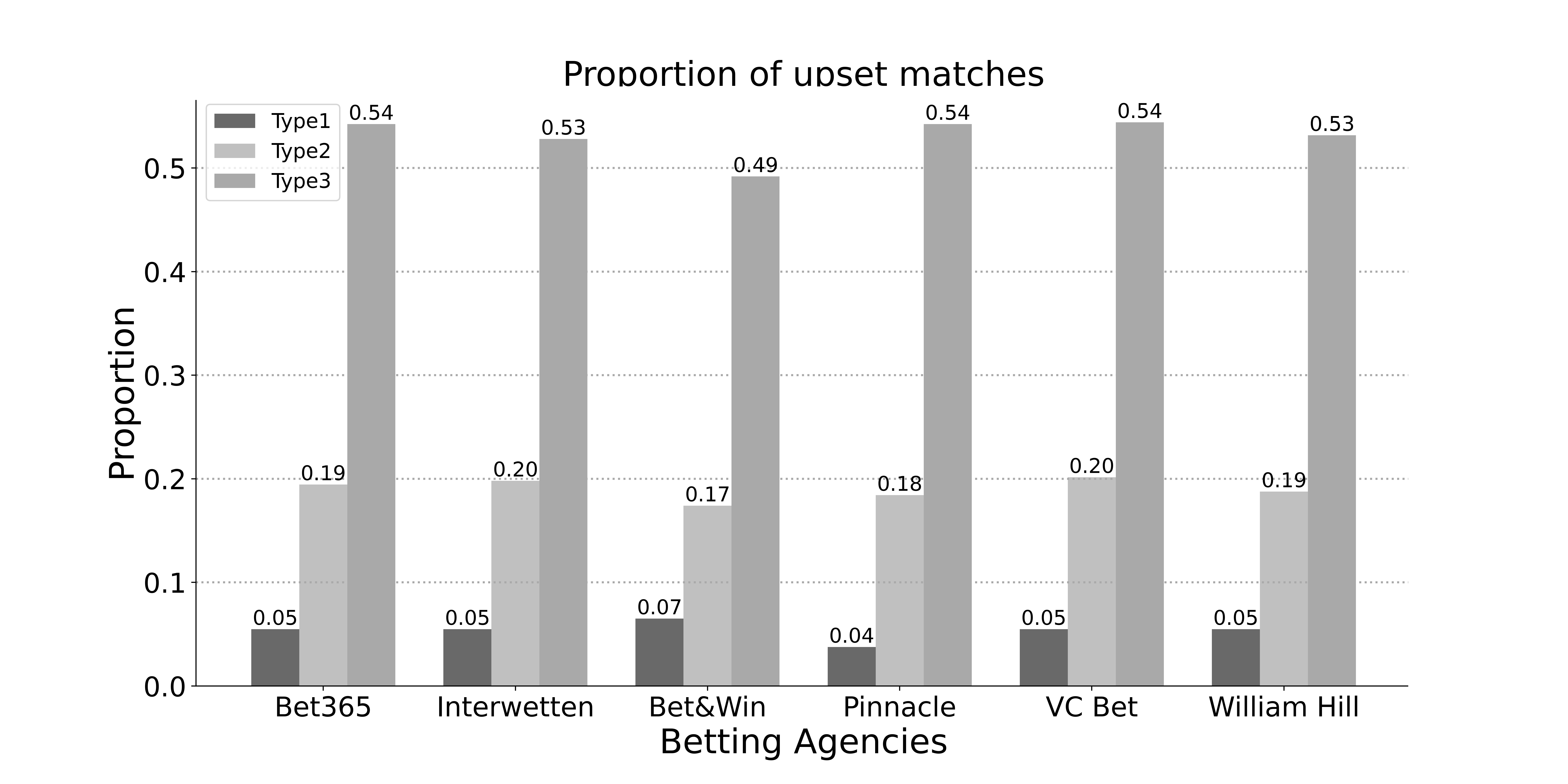}
\caption{Percentage of upset in each type of match}
\label{fig:Percentage of upset in each type of match}
\end{figure}

Although bookmakers can adjust the odds based on their own predictions, upsets often happen in football matches. It is not always straightforward to precisely define which matches are upsets based on the available data. This study identifies matches as an upset when the study's prediction was correct and the bookmaker not only predicted incorrectly, but also set the highest odds on the actual outcome of the match. See the next sub-section for the predicted results. The percentage of upsets for each type is shown in Figure~\ref{fig:Percentage of upset in each type of match}.

\begin{figure}[htb]
\centering
\includegraphics[width=0.6\textwidth]{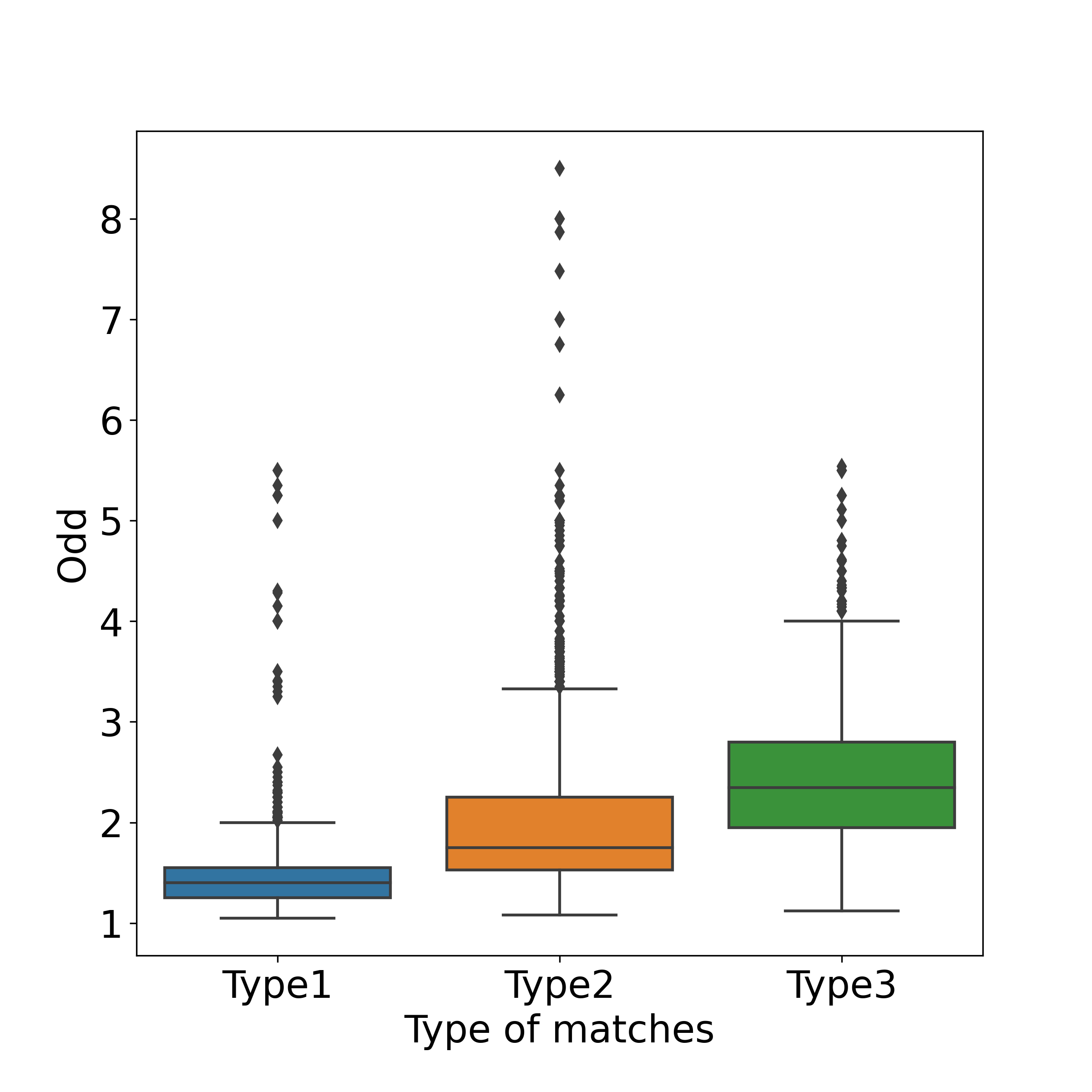}
\caption{Distribution of odds set by the six bookmakers on the predicted results of this study}
\label{fig:Distribution of odds set by the six bookmakers on the predicted results of this study}
\end{figure}

Betting agencies are poor predictors of Type 3 matches compared to the other two types. Upsets are very unlikely in the Type 1 matches, but they become a regular occurrence in the Type 3 matches. Figure~\ref{fig:Distribution of odds set by the six bookmakers on the predicted results of this study} aggregates the spread of the odds across all six betting agencies in this study. According to this figure, on average, bookmakers set significantly lower odds on results in Type 1 matches than in the other two categories. 

\subsection{Result prediction}
The prediction accuracy, precision, recall and average ranking of each algorithm across the prediction process for Type 1, Type 2 and Type 3 matches are all shown in Table~\ref{Prediction results from various machine learning algorithms in the Type 1 match}, Table~\ref{Prediction results from various machine learning algorithms in the Type 2 match} and Table~\ref{Prediction results from various machine learning algorithms in the Type 3 match} respectively. The algorithms are rank-ordered from best-performing onwards. The average ranking is likewise calculated by the accuracy of each algorithm across all different forecasting windows. Meanwhile, the accuracies of all the algorithms across the different Types are contrasted with the accuracies attained across all matches without taking the Kelly Index into account. These results can be seen in Table~\ref{Prediction results from various machine learning algorithms in the Type all matches}.

\begin{table}[]
\centering
\caption{\label{Prediction results from various machine learning algorithms in the Type 1 match}Prediction results from various machine learning algorithms in the Type 1 match}
\fontsize{7pt}{9pt} 
\selectfont
\begin{tabular}{@{}
>{\columncolor[HTML]{FFFFFF}}l 
>{\columncolor[HTML]{FFFFFF}}c 
>{\columncolor[HTML]{FFFFFF}}c 
>{\columncolor[HTML]{FFFFFF}}c 
>{\columncolor[HTML]{FFFFFF}}c 
>{\columncolor[HTML]{FFFFFF}}c @{}}
\toprule
\textbf{Algorithm} & \textbf{Accuracy} & \textbf{Precision} & \textbf{Recall} & \textbf{F1\_Score} & \textbf{Rank} \\ \midrule
CatBoost           & 70.0\% & 66.6\% & 54.8\% & 63.6\% & 1.1 \\
Logistic Regression & 67.9\% & 64.8\% & 53.6\% & 62.3\% & 3.1 \\
Random Forest       & 68.9\% & 64.3\% & 54.9\% & 63.4\% & 3.2 \\
Voting              & 66.6\% & 59.4\% & 52.4\% & 61.7\% & 4.3 \\
KNN                 & 62.1\% & 60.6\% & 52.8\% & 61.3\% & 4.5 \\
Gradient Boosting   & 62.5\% & 59.6\% & 51.7\% & 60.8\% & 4.8 \\
Stacking            & 65.5\% & 54.1\% & 50.9\% & 59.1\% & 5.0 \\
Decision Tree       & 60.4\% & 56.7\% & 48.9\% & 58.2\% & 5.1 \\
XGBoost            & 63.8\% & 57.1\% & 51.0\% & 59.6\% & 5.3 \\
Baseline 1               & 47.1\% & 33.5\% & 31.5\% & 37.4\% & 8.8 \\
Baseline 2        & 30.4\% & 36.9\% & 29.9\% & 32.1\% & 9.8 \\ \bottomrule
\end{tabular}
\end{table}

\begin{table}[]
\centering
\caption{\label{Prediction results from various machine learning algorithms in the Type 2 match}Prediction results from various machine learning algorithms in the Type 2 match}
\fontsize{7pt}{9pt} 
\selectfont
\begin{tabular}{@{}
>{\columncolor[HTML]{FFFFFF}}l 
>{\columncolor[HTML]{FFFFFF}}c 
>{\columncolor[HTML]{FFFFFF}}c 
>{\columncolor[HTML]{FFFFFF}}c 
>{\columncolor[HTML]{FFFFFF}}c 
>{\columncolor[HTML]{FFFFFF}}c @{}}
\toprule
\textbf{Algorithm} & \textbf{Accuracy} & \textbf{Precision} & \textbf{Recall} & \textbf{F1\_Score} & \textbf{Rank} \\ \midrule
CatBoost           & 49.7\% & 38.8\% & 42.0\% & 43.3\% & 2.8 \\
Logistic Regression & 51.0\% & 49.3\% & 45.4\% & 48.8\% & 3.1 \\
Random Forest       & 52.7\% & 50.9\% & 46.5\% & 49.2\% & 3.4 \\
Decision Tree       & 45.9\% & 45.3\% & 42.6\% & 45.6\% & 3.7 \\
Stacking            & 49.3\% & 44.5\% & 42.1\% & 43.8\% & 4.3 \\
Voting              & 47.9\% & 44.5\% & 42.7\% & 45.7\% & 4.9 \\
XGBoost            & 48.6\% & 46.0\% & 43.6\% & 46.7\% & 5.0 \\
KNN                 & 39.7\% & 40.1\% & 37.5\% & 39.9\% & 5.9 \\
Gradient Boosting   & 40.8\% & 39.7\% & 37.6\% & 40.0\% & 6.2 \\
Baseline 1               & 40.1\% & 29.0\% & 32.3\% & 31.7\% & 6.9 \\
Baseline 2        & 31.5\% & 33.5\% & 31.3\% & 32.1\% & 8.7 \\ \bottomrule
\end{tabular}
\end{table}

\begin{table}[]
\centering
\caption{\label{Prediction results from various machine learning algorithms in the Type 3 match}Prediction results from various machine learning algorithms in the Type 3 match}
\fontsize{7pt}{9pt} 
\selectfont
\begin{tabular}{@{}
>{\columncolor[HTML]{FFFFFF}}l 
>{\columncolor[HTML]{FFFFFF}}c 
>{\columncolor[HTML]{FFFFFF}}c 
>{\columncolor[HTML]{FFFFFF}}c 
>{\columncolor[HTML]{FFFFFF}}c 
>{\columncolor[HTML]{FFFFFF}}c @{}}
\toprule
\textbf{Algorithm} & \textbf{Accuracy} & \textbf{Precision} & \textbf{Recall} & \textbf{F1\_Score} & \textbf{Rank} \\ \midrule
Decision Tree       & 40.0\% & 38.5\% & 37.5\% & 38.5\% & 3.9 \\
CatBoost           & 40.0\% & 37.6\% & 36.3\% & 35.2\% & 4.0 \\
Random Forest       & 41.1\% & 39.4\% & 38.4\% & 39.1\% & 4.2 \\
Logistic Regression & 40.7\% & 38.8\% & 38.2\% & 39.1\% & 4.5 \\
Voting              & 40.0\% & 39.1\% & 37.7\% & 38.5\% & 4.6 \\
Gradient Boosting   & 37.7\% & 38.0\% & 37.0\% & 37.8\% & 5.0 \\
Stacking            & 39.1\% & 35.5\% & 35.9\% & 36.2\% & 5.2 \\
XGBoost            & 39.6\% & 38.8\% & 37.9\% & 39.0\% & 5.4 \\
KNN                 & 35.3\% & 34.9\% & 33.9\% & 34.8\% & 5.8 \\
Baseline 2        & 37.3\% & 38.3\% & 37.0\% & 37.6\% & 6.2 \\
Baseline 1               & 36.4\% & 26.3\% & 33.4\% & 26.4\% & 6.4 \\ \bottomrule
\end{tabular}
\end{table}

\begin{table}[]
\centering
\caption{\label{Prediction results from various machine learning algorithms in the Type all matches}Prediction results from various machine learning algorithms in the all matches}
\fontsize{7pt}{9pt} 
\selectfont
\begin{tabular}{@{}
>{\columncolor[HTML]{FFFFFF}}l 
>{\columncolor[HTML]{FFFFFF}}c 
>{\columncolor[HTML]{FFFFFF}}c 
>{\columncolor[HTML]{FFFFFF}}c 
>{\columncolor[HTML]{FFFFFF}}c 
>{\columncolor[HTML]{FFFFFF}}c @{}}
\toprule
\textbf{Algorithm} & \textbf{Accuracy} & \textbf{Precision} & \textbf{Recall} & \textbf{F1\_Score} & \textbf{Rank} \\ \midrule
CatBoost           & 51.9\% & 45.7\% & 44.7\% & 45.3\% & 2.6 \\
Logistic Regression & 50.6\% & 44.0\% & 43.5\% & 45.2\% & 3.2 \\
Random Forest       & 52.0\% & 47.7\% & 44.7\% & 45.7\% & 3.4 \\
Stacking            & 51.3\% & 46.7\% & 44.5\% & 45.9\% & 3.6 \\
Decision Tree       & 50.2\% & 46.6\% & 44.6\% & 46.2\% & 3.8 \\
Voting              & 49.9\% & 46.7\% & 44.5\% & 47.0\% & 4.6 \\
Gradient Boosting   & 47.1\% & 45.8\% & 43.3\% & 46.3\% & 5.0 \\
XGBoost            & 48.4\% & 45.7\% & 43.5\% & 46.5\% & 5.7 \\
KNN                 & 45.4\% & 44.4\% & 42.0\% & 44.8\% & 5.7 \\
Baseline 1               & 40.5\% & 28.4\% & 32.8\% & 29.7\% & 8.0 \\
Baseline 2        & 31.9\% & 33.2\% & 32.1\% & 32.2\% & 9.2 \\ \bottomrule
\end{tabular}
\end{table}

The model offering the greatest predictive performance for the Type 1 matches across all forecasting windows was produced by  CatBoost, with an accuracy of 70\%. This performance was significantly better than that of the baseline methods which scored well below 50\%.

The best model in the Type 2 matches was generated by Random Forest, although Catboost ranked as the winner across all the forecasted windows. Random Forest achieved 53\% which represents a significant reduction in accuracy compared to Type 1 models. However, the best-performing models under the Type 2 setting still clearly outperformed baseline predictions.

The accuracy of Type 3 accuracies continued the decrease down to 41\%, which was the best accuracy registered by  Random Forest. Though overall, both the Decision Tree and CatBoost performed more consistently across all the forecasting windows. While the best-performing models' efficacy significantly reduced for Type 3 matches, they still marginally contributed more value than baseline models.
Unexpectedly, stacking and voting methods were consistently outperformed by other algorithms which is in contrast to the performance of these algorithms in other studies covering competitive sports.

The results from the previous tables are depicted in Figure~\ref{fig:As the extended window progresses, the trend in prediction accuracy of the algorithms with the best predictive ability for different types of matches} highlighting the accuracy of each of the best-performing algorithms across time for each season, and Type of matches. The dashed lines in the figure represent interpolations in cases where no matches were predicted for a given Type when matches meeting the Kelly Index criteria did not exist. The figure demonstrates that Type 1 predictions consistently outperformed other Types.
.
Taken altogether, the results demonstrate the  efficacy of the Kelly Index to reduce an element of uncertainty from the predictive problem and thereby increase the overall predictive accuracy.

\begin{figure}[hbt]
\centering
\includegraphics[width=1.0\textwidth]{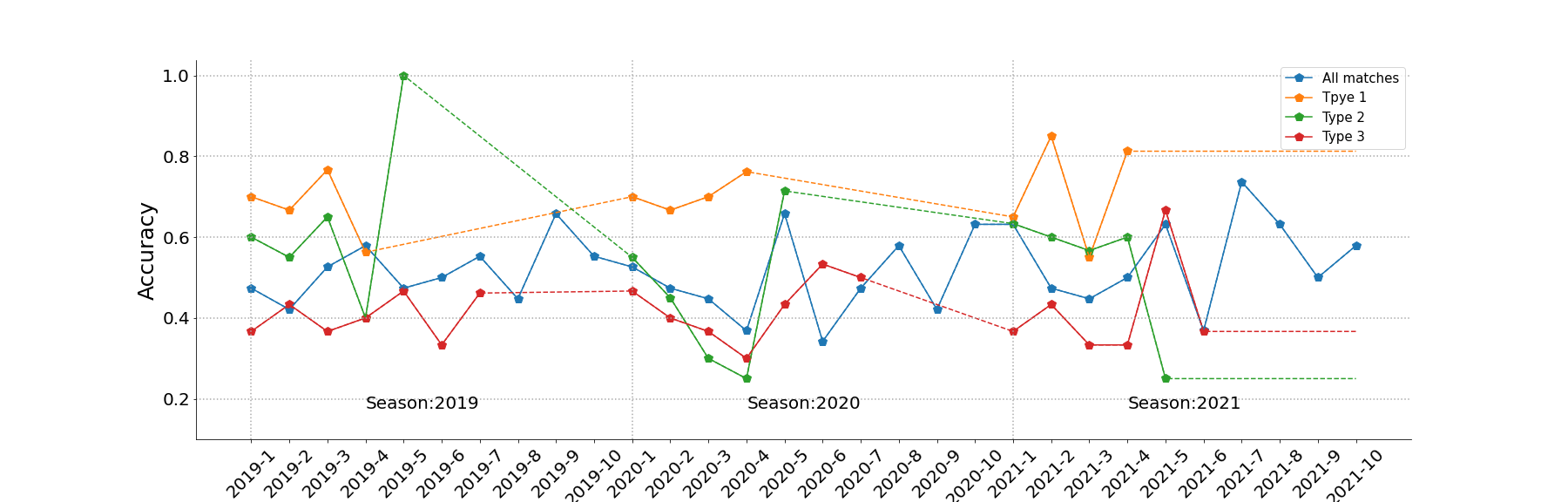}
\caption{As the extended window progresses, the trend in prediction accuracy of the algorithms with the best predictive ability for different types of matches.}
\label{fig:As the extended window progresses, the trend in prediction accuracy of the algorithms with the best predictive ability for different types of matches}
\end{figure}

\begin{figure}[H]
\centering
\includegraphics[width=1.0\textwidth]{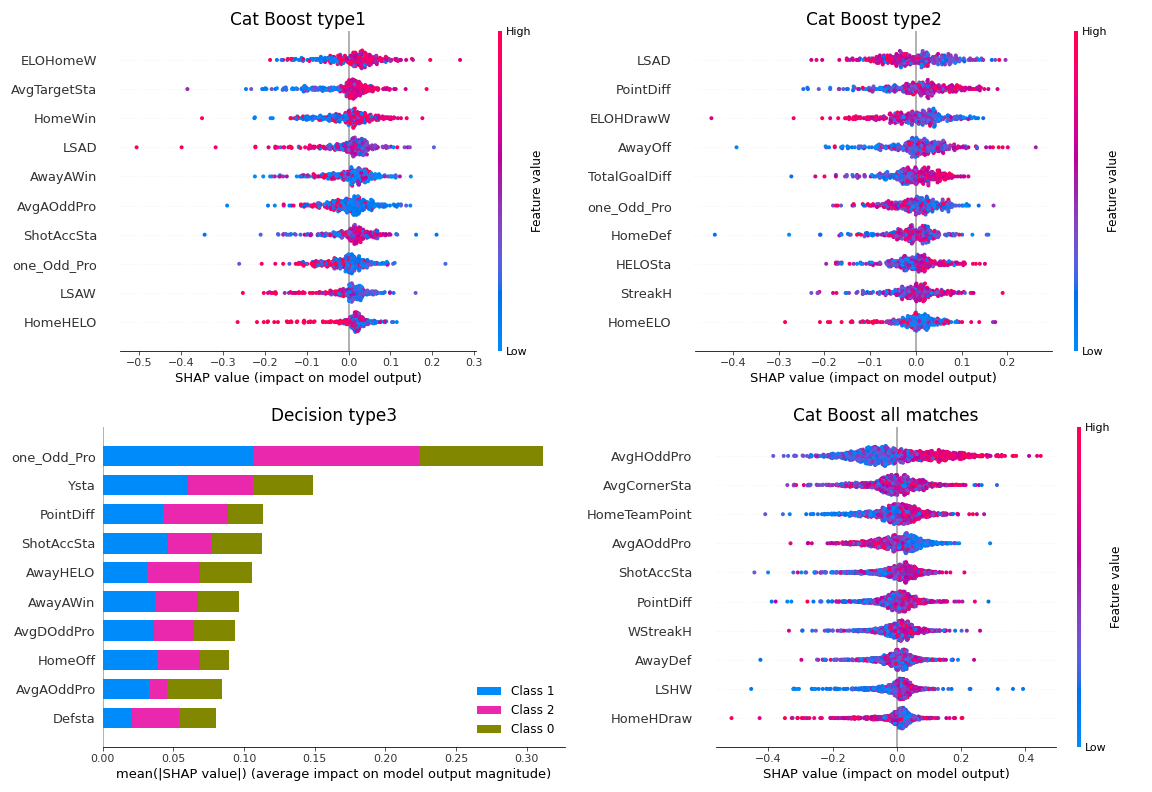}
\caption{Scatterplot of SHAP values for the top 10 features of importance when being predicted for the three different types of matches and for all matches.}
\label{fig:Scatterplot of SHAP values for the top 10 features of importance when being predicted for the three different types of matches and for all matches.}
\end{figure}

The next analysis considers the internals of the predictive moedls and uses SHAP in order to understand the impact of the key features on the best-performing algorithms based on their ranks, across each of the Types of matches. The feature importances and their effects are shown in Figure~\ref{fig:Scatterplot of SHAP values for the top 10 features of importance when being predicted for the three different types of matches and for all matches.}.  

For Type 1 matches, the top most influential features are based on the team's historical match statistics based on their home-win record. These are matches where the home team's advantage exists and it is easier to determine whether the match will be one-sided based on the features of the home team. The rarety of draw results of such matches also leads to a higher prediction accuracy. 

In the cases of Type 2 matches, we can see that as the prediction becomes more difficult, historical data on both teams becomes less important in and is replaced by odds data provided by bookmakers before the match.

Meanwhile, in predicting the Type 3 matches, historical data appears to be inconsequential. The features based on bookmakers' prediction estimates  still have the advantage in predicting these matches. The already low accuracy accuracy of the Type 3 matches still depends on the results of the bookmakers' forecasts. The odds in this case  have an almost direct influence on the prediction of uncategorised matches.

\subsection{Betting returns}
This section compares traditional betting methods in football betting with the returns of classified matches betting to examine if there is a way to maximise returns or minimise risk. We first begin by highlighting the distribution in confidence probabilities of the machine learning algorithms across different Types as this will form a key factor when comparing betting strategies. 

Table \ref{The distribution of confidence in the predicted results of each type across seasons in different ranges} shows the distribution of confidence in the predicted outcomes for the three types of matches. Even for the most predictable matches, machine learning has difficulty predicting a match's result with confidence. Surprisingly, the proportion of these predictions that machine learning considered extremely confident was similar for the easy-to-predict matches and the hard-to-predict matches, and these matches were rarely present in both types of matches. 

\begin{table}
\centering
\caption{\label{The distribution of confidence in the predicted results of each type across seasons in different ranges}The distribution of confidence in the predicted results of each type across seasons in different ranges}
\fontsize{7pt}{9pt} 
\selectfont
\begin{tabular}{lccccccc}
\toprule
\multirow{2}{*}{Type}  & \multicolumn{1}{l}{\multirow{2}{*}{Season}} & \multicolumn{5}{c}{Range of confidence}                                                      & \multicolumn{1}{l}{\multirow{2}{*}{Match number}}  \\
                       & \multicolumn{1}{l}{}                        & 0.7-1.0          & 0.6-0.7          & 0.5-0.6          & 0.4-0.5          & 0.33-0.4         & \multicolumn{1}{l}{}                               \\\midrule
\multirow{3}{*}{Type1} & \vcell{2019}                                & \vcell{8.5\%}    & \vcell{10.4\%}   & \vcell{15.09\%}  & \vcell{55.7\%}   & \vcell{10.4\%}   & \vcell{106}                                        \\[-\rowheight]
                       & \printcellmiddle                            & \printcellmiddle & \printcellmiddle & \printcellmiddle & \printcellmiddle & \printcellmiddle & \printcellmiddle                                   \\
                       & 2020                                        & 0.0\%            & 0.0\%            & 4.50\%           & 42.3\%           & 53.2\%           & 111                                                \\
                       & 2021                                        & 2.6\%            & 25.0\%           & 27.63\%          & 36.8\%           & 7.9\%            & 76                                                 \\
\multirow{3}{*}{Type2} & \vcell{2019}                                & \vcell{17.3\%}   & \vcell{28.4\%}   & \vcell{13.58\%}  & \vcell{25.9\%}   & \vcell{14.8\%}   & \vcell{81}                                         \\[-\rowheight]
                       & \printcellmiddle                            & \printcellmiddle & \printcellmiddle & \printcellmiddle & \printcellmiddle & \printcellmiddle & \printcellmiddle                                   \\
                       & 2020                                        & 5.8\%            & 10.3\%           & 33.33\%          & 41.4\%           & 9.2\%            & 87                                                 \\
                       & 2021                                        & 11.3\%           & 33.1\%           & 18.55\%          & 24.2\%           & 12.9\%           & 124                                                \\
\multirow{3}{*}{Type3} & \vcell{2019}                                & \vcell{6.2\%}    & \vcell{6.2\%}    & \vcell{19.69\%}  & \vcell{48.7\%}   & \vcell{19.2\%}   & \vcell{193}                                        \\[-\rowheight]
                       & \printcellmiddle                            & \printcellmiddle & \printcellmiddle & \printcellmiddle & \printcellmiddle & \printcellmiddle & \printcellmiddle                                   \\
                       & 2020                                        & 2.2\%            & 8.2\%            & 21.98\%          & 47.8\%           & 19.8\%           & 182                                                \\
                       & 2021                                        & 2.8\%            & 8.9\%            & 21.11\%          & 47.2\%           & 20.0\%           & 180              \\
                       \midrule
\end{tabular}
\end{table}

We next construct a simulated betting scenario in order to determine the return on investment(ROI) based on several strategies. We hypothetically invest \$1 on each match and receive corresponding profits according to the odds for a correct prediction, and a loss of the investment of \$1 for a wrong prediction. The ROI is calculated by dividing the investment profit by the total amount invested. A positive or negative ROI is used to determine whether an investment strategy is profitable or not.

Firstly, a set of benchmark ROIs are prepared for this study and are shown in Table~\ref{All invested in one type of return}. These are returns based on the model with the highest prediction accuracy across each of the Kelly index categories as well as the baseline which includes all matches, investing in the six bookmakers. A naive approach indicates marginal returns on two cases only, with higher losses across all other scenarios. 

\begin{table}[]
\centering
\caption{\label{All invested in one type of return}Return on total investment by type of competition}
\fontsize{7pt}{9pt} 
\selectfont
\begin{tabular}{@{}
>{\columncolor[HTML]{FFFFFF}}l 
>{\columncolor[HTML]{FFFFFF}}c 
>{\columncolor[HTML]{FFFFFF}}c 
>{\columncolor[HTML]{FFFFFF}}c 
>{\columncolor[HTML]{FFFFFF}}c @{}}
\toprule
\textbf{}             & \textbf{Type1} & \textbf{Type2} & \textbf{Type3} & \textbf{Baseline} \\ \midrule
\textbf{Match number}   & 292            & 293            & 555            & 1140            \\ \hline
\textbf{Bet365}       & -1.1\%         & -5.9\%         & -3.8\%         & -6.7\%            \\
\textbf{Interwetten}  & 0.9\%          & -4.8\%         & -3.4\%         & -5.4\%            \\
\textbf{Bet\&Win}     & -0.2\%         & -5.5\%         & -3.8\%         & -6.2\%            \\
\textbf{Pinnacle}     & 1.4\%          & -3.5\%         & -1.1\%         & -4.1\%            \\
\textbf{VC Bet}       & -0.7\%         & -6.0\%         & -3.3\%         & -6.3\%            \\
\textbf{William Hill} & -0.5\%         & -5.6\%         & -3.6\%         & -6.1\%            \\ \bottomrule
\end{tabular}
\end{table}

The next step in the experiments was to calibrate the investment strategy based on the confidence level of the predictive model in the outcome of each match. Thus, a bet would be placed only if the predictive model's confidence level met or exceeded a predefined threshold. Several Thresholds were chosen and examined for returns. For each Type level, the best-performing model was selected. Odds from Pinnacle were chosen.  

\begin{figure}[H]
\centering
\includegraphics[width=1.0\textwidth]{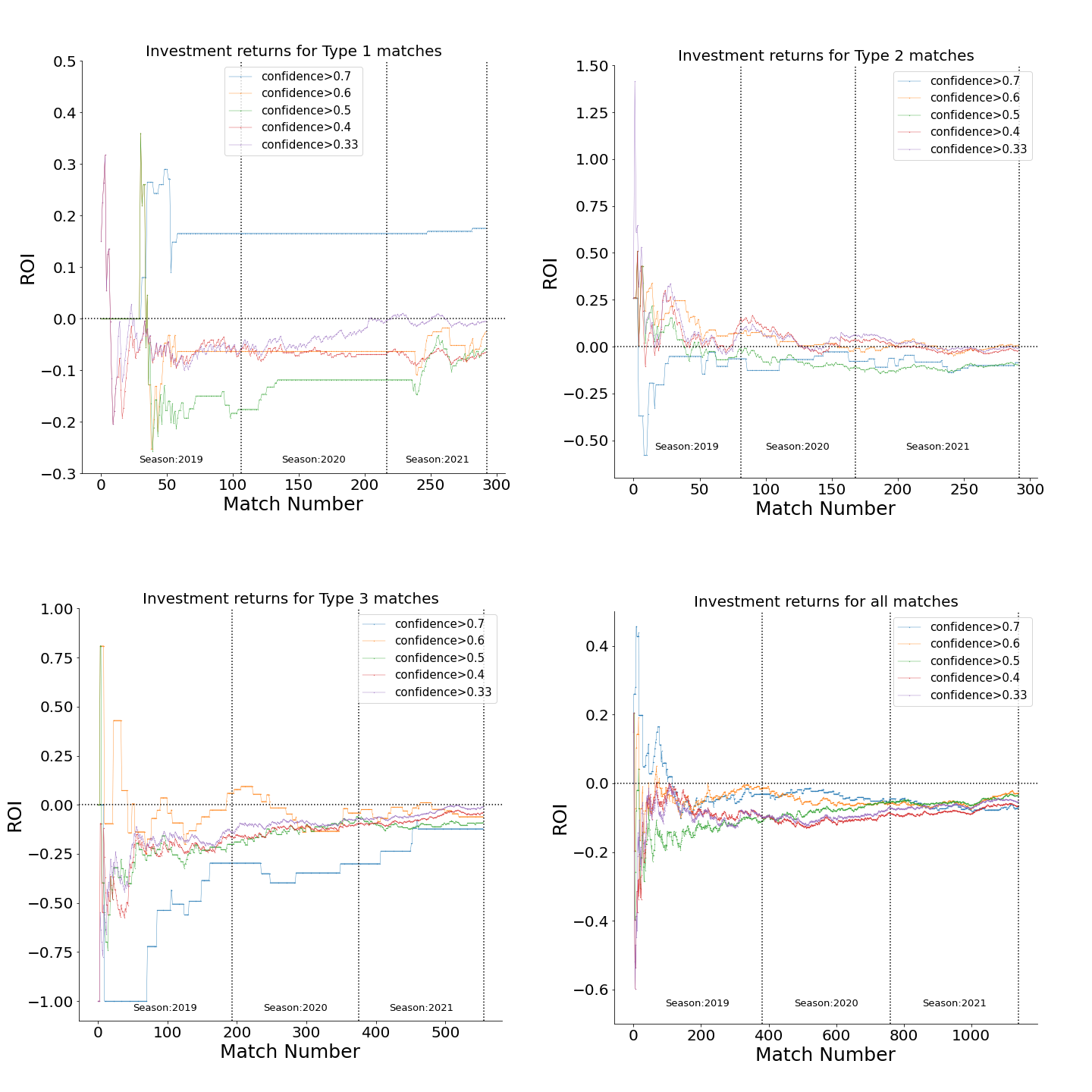}
\caption{A graph of the return on investment as the match progresses, based on how confident the machine learning method is in predicting the result to determine whether to invest in the match.}
\label{fig:Trend in returns}
\end{figure}

The results of the experiments are shown in Figure \ref{fig:Trend in returns} highlighting the ROI across time and seasons. Over time, all strategies based both on Type and models' confidence level conclude with a negative ROI. The exception to this is found in a single strategy based on Type 1 matches and a confidence threshold of 70\%. The final ROI using this strategy is ~17\%.

\section{Conclusions and future work}\label{sec13}

This study has considered the problem of predicting the outcomes of football matches using the Premier League match data from 2019-2021 seasons.  The proposed strategy used the Kelly Index to first categorise the matches into three groups, with each one representing different levels of uncertainty, or predictability. A range of machine learning algorithms were explored for predicting the outcomes of the matches from each category in order to determine the utility of the process of decomposing the predictive problem into sub-tasks. This paper devised a range of new features previously unexplored as well as machine learning algorithms not investigated in this domain. The study found that ensemble-based algorithms outperformed all other approaches including the benchmark approaches, while the models produced competitive results with prior works.

The paper also validates the proposed approaches by benchmarking them against bookmaker odds in order to determine with strategies are able to return a profit on investment. A method was developed that combines both the Kelly Index together with predictive confidence thresholds and investigated. The findings indicate that a strategy comprised of a combination of focusing on easy-to-predict matches that have a high predictive confidence level from machine learning models can return a profit over the long term.  is a non-negligible part of the match data for machine learning. Moreover, the available features cannot explain the reasons for the match result.

%\bibliographystyle{plainnat}
%\bibliography{references.bib}% common bib file

\urlstyle{same}

\appendix

\section{Additional Accuracy Results}

\begin{figure}[H]
\centering
\includegraphics[width=0.8\textwidth]{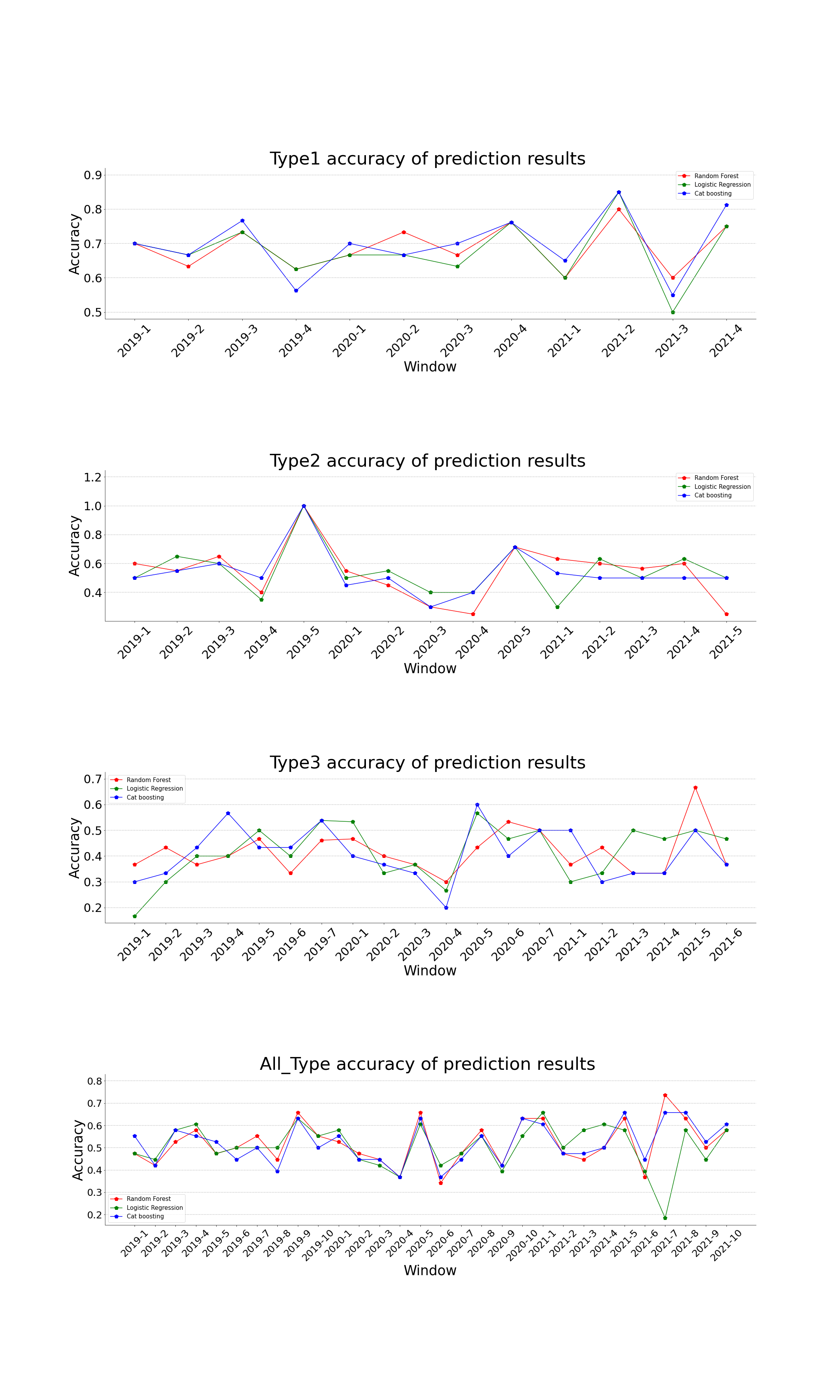}
\caption{Prediction accuracy per window for three machine learning algorithms that produce excellent prediction results}
\label{fig:Prediction accuracy per window for three machine learning algorithms that produce excellent prediction results}
\end{figure}

The best three performing algorithms from the above tables are selected and their detailed accuracies are depicted in Figure~\ref{fig:Prediction accuracy per window for three machine learning algorithms that produce excellent prediction results} which shows their relative accuracies across time.  A pattern can be detected where the divergence in accuracies becomes increasingly pronounced between different algorithms as the Type of matches being predicted increases, indicating a higher uncertainty.

\begin{figure}[H]
\centering
\includegraphics[width=0.9\textwidth]{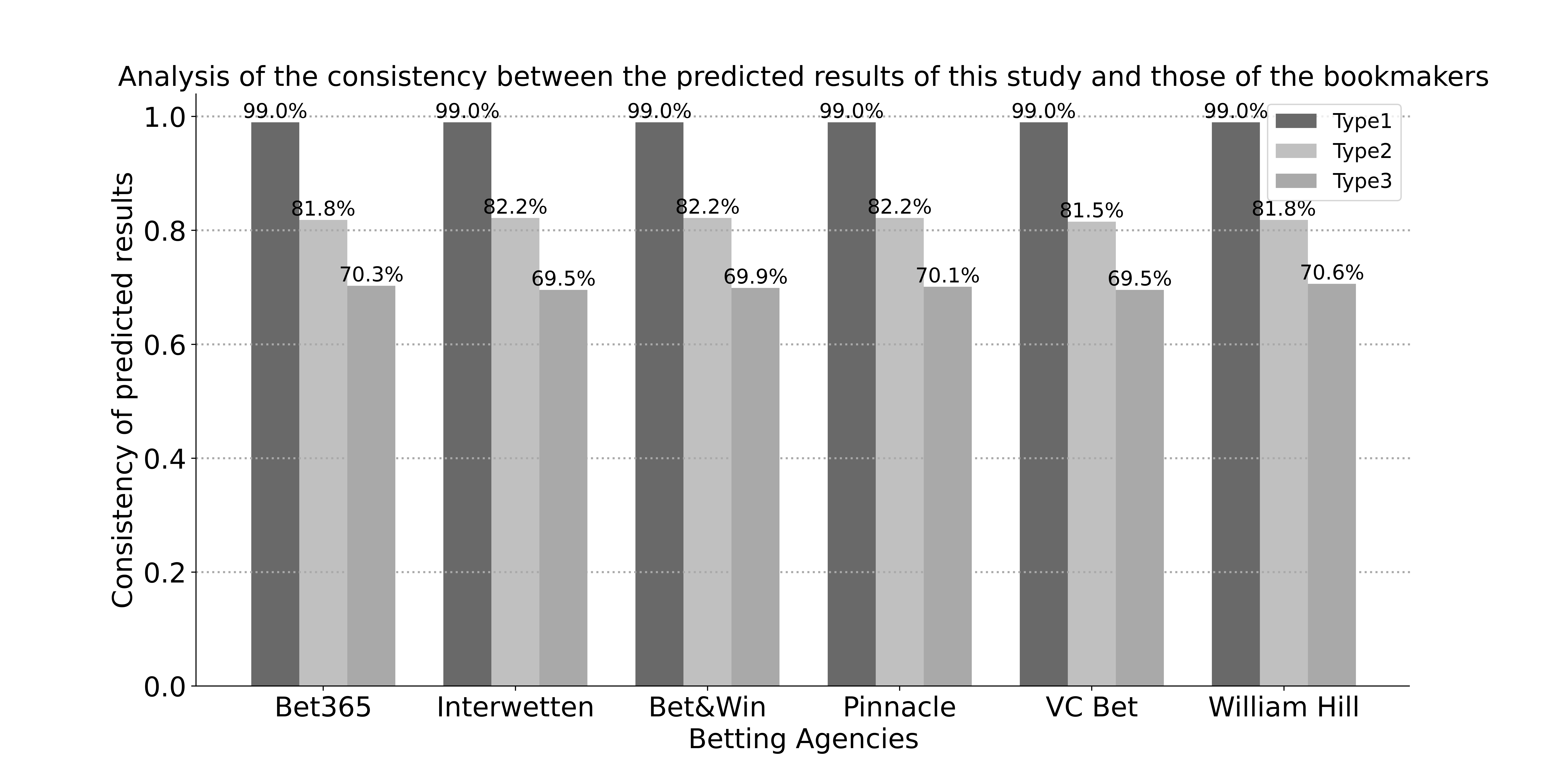}
\caption{The consistency of the different bookmakers' predictions in terms of results with those of this study.}
\label{fig:The consistency of the different bookmakers' predictions in terms of results with those of this study.}
\end{figure}

%%%%%%%%%%%%%
%%%%%%%%%%%%% consistency
%%%%%%%%%%%%%
Figure~\ref{fig:The consistency of the different bookmakers' predictions in terms of results with those of this study.} shows whether the predictions of this study are the same as those of the bookmakers. The vertical axis shows the proportion of all matches where the two predictions are the same. Both are almost identical in their predictions for Type 1 matches. However, as the difficulty of predicting matches increases, the difference between the two increases, although they are similar in terms of prediction accuracy. There is no way of knowing what algorithms and features the bookmakers used in their predictions. However, in terms of results, a model using only historical match result data can achieve the predictive power of a bookmaker's forecasting model. This also demonstrates that it is practical and meaningful to use the Kelly Index to classify football matches.

There is no one fixed way to predict the result of a match. There is no pattern between the features used when predicting from different windows. The available features do not explain how the results of football matches are generated. This is one of the reasons why pre-match prediction has hit an upper accuracy ceiling.

The high fluctuation in accuracy over a short period of time is what makes the game of football so attractive as a high-intensity competition. This explains the necessity of using the extended window method or the rolling window method when predicting competitive events that occur in the same chronological order as a football match. When predicting matches that occur over a short period of time, such as a single season, the changes in competitive strategy, team personnel changes and changes in player mentality that arise from advancing time are minimised. It is possible for a machine learning algorithm to produce a model that is overfitting for the entire league's history of matches but predicts a particular season's matches well. However, such models have no practical implications for future matches.

\end{document}